\documentclass[runningheads]{llncs}
\usepackage{graphicx}
\usepackage{hyperref}
\hypersetup{
    colorlinks,
    citecolor=green
}
\usepackage{tikz}
\usepackage{comment}
\usepackage{amsmath,amssymb} 
\usepackage{color}
\usepackage{graphicx}
\usepackage{amsmath}
\usepackage{amssymb}
\usepackage{booktabs}
\usepackage{nicefrac}
\usepackage{multicol}
\usepackage{multirow}
\usepackage{xspace}
\usepackage{array}
\newcolumntype{x}[1]{>{\centering\arraybackslash\hspace{0pt}}p{#1}}
\usepackage[accsupp]{axessibility}  

\newcommand{\eg}{\textit{e.g.~\xspace}}
\newcommand{\ie}{\textit{i.e.~\xspace}}
\newcommand{\etal}{\textit{et al.~\xspace}}
\DeclareMathOperator*{\argmax}{arg\,max}
\DeclareMathOperator*{\argmin}{arg\,min}

\begin{document}
\pagestyle{headings}
\mainmatter
\def\ECCVSubNumber{5556}  

\title{On the Robustness of Quality Measures for GANs} 

\titlerunning{ECCV-22 submission ID \ECCVSubNumber} 
\authorrunning{ECCV-22 submission ID \ECCVSubNumber} 
\author{Anonymous ECCV submission}
\institute{Paper ID \ECCVSubNumber}
\titlerunning{On the Robustness of Quality Measures for GANs}
%
\author{Motasem Alfarra\inst{1} \and
Juan C. P\'erez\inst{1} \and
Anna Fr\"{u}hst\"{u}ck\inst{1}
\and Philip H. S. Torr\inst{2}\and\\
 Peter Wonka\inst{1} \and
Bernard Ghanem\inst{1}}
\authorrunning{Motasem Alfarra \textit{et al.}}
%
\institute{King Abdullah University of Science and Tehchnology (KAUST), Saudi Arabia \and
University of Oxford, United Kingdom\\
\email{motasem.alfarra@kaust.edu.sa}\\
}
\maketitle

\begin{abstract}
    This work evaluates the robustness of quality measures of generative models such as Inception Score (IS) and Fréchet Inception Distance (FID). 
    Analogous to the vulnerability of deep models against a variety of adversarial attacks, we show that such metrics can also be manipulated by additive pixel perturbations.
    Our experiments indicate that one can generate a distribution of images with very high scores but low perceptual quality.
    Conversely, one can optimize for small imperceptible perturbations that, when added to real world images, deteriorate their scores.
    We further extend our evaluation to generative models themselves, including the state of the art network StyleGANv2.
    We show the vulnerability of both the generative model and the FID against additive perturbations in the latent space.
    Finally, we show that the FID can be robustified by simply replacing the standard Inception with a robust Inception. 
    We validate the effectiveness of the robustified metric through extensive experiments, showing it is more robust against manipulation.\footnote{Code: \href{https://github.com/MotasemAlfarra/R-FID-Robustness-of-Quality-Measures-for-GANs}{https://github.com/R-FID-Robustness-of-Quality-Measures-for-GANs}}
\end{abstract}

\section{Introduction}

\begin{figure*}
    \centering
    \includegraphics[width=\textwidth]{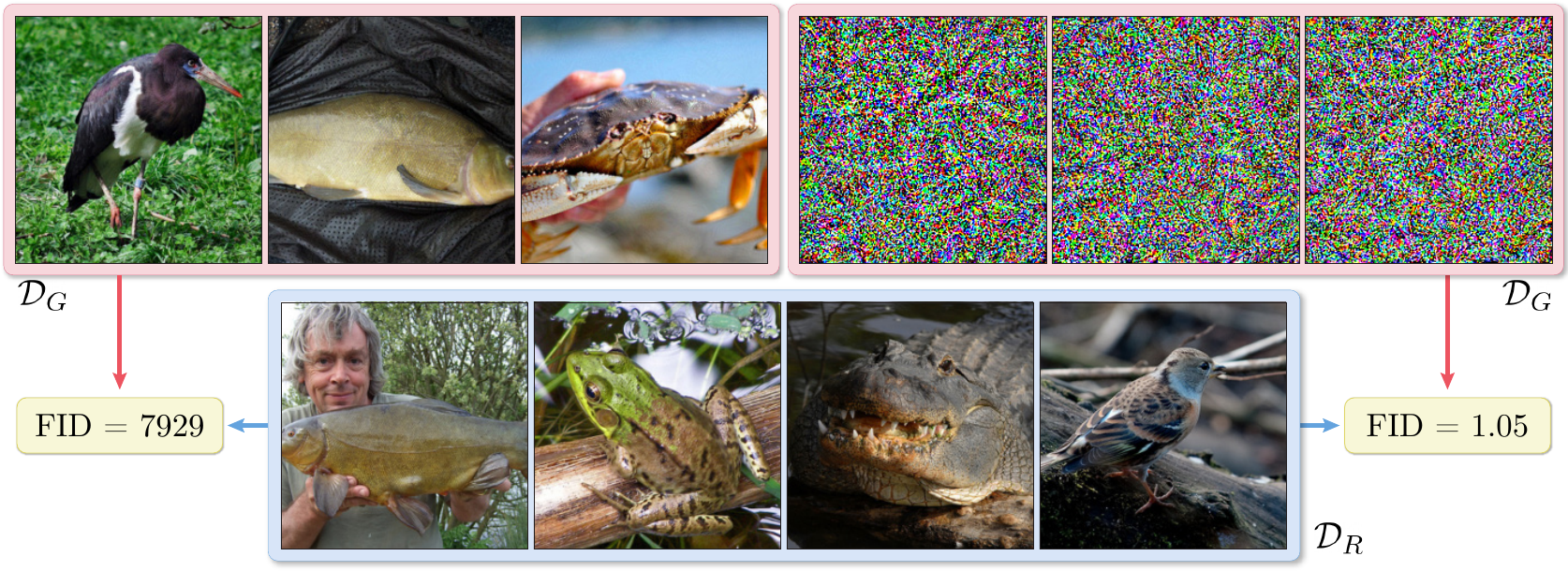}
    \caption{
    \textbf{Does the Fréchet Inception Distance (FID) accurately measure the distances between image distributions?}
    We generate datasets that demonstrate the unreliability of FID in judging perceptual (dis)similarities between image distributions.
    The top left box shows a sample of a dataset constructed by introducing imperceptible noise to each ImageNet image. 
    Despite the remarkable visual similarity between this dataset and ImageNet (bottom box), an extremely large  FID (almost 8000) between these two datasets  showcases FID's failure to capture perceptual similarities.
    On the other hand, a remarkably low FID (almost 1.0) between a dataset of random noise images (samples shown in the top right box) and ImageNet illustrates FID's failure to capture perceptual dissimilarities. 
    }
    \label{fig:pull}
\end{figure*}

Deep Neural Networks (DNNs) are vulnerable to small imperceptible perturbations known as adversarial attacks. 
For example, while two inputs $x$ and $(x+\delta)$ can be visually indistinguishable to humans, a classifier $f$ can output two different predictions. 
To address this deficiency in DNNs, adversarial attacks~\cite{goodfellow2015explaining,croce2020reliable} and defenses~\cite{madry2018towards,mart} have prominently emerged as active areas of research. 
Starting from image classification~\cite{awp}, researchers also assessed the robustness of DNNs for other tasks, such as segmentation~\cite{arnab2018robustness}, object detection~\cite{zhao2019seeing}, and point cloud classification~\cite{liu2021pointguard}.
While this lack of robustness questions the reliability of DNNs and hinders their deployment in the real world, DNNs are still widely used to evaluate performance in other computer vision tasks, such as that of generation.

Metrics in use for assessing generative models in general, and Generative Adversarial Networks (GANs)~\cite{goodfellow2014generative} in particular, are of utmost importance in the literature.
This is because such metrics are widely used to establish the superiority of a generative model over others, hence guiding which GAN should be deployed in real world.
Consequently, such metrics are expected to be not only useful in providing informative statistics about the distribution of generated images, but also reliable and robust.
In this work, we investigate the robustness of metrics used to assess GANs. 
We first identify two interesting observations that are unique to this context.
First, current GAN metrics are built on pretrained classification DNNs that are nominally trained (\ie trained on clean images only). 
A popular DNN of choice is the Inception model~\cite{szegedy2016rethinking}, on which the Inception Score (IS)~\cite{salimans2016improved} and Fréchet Inception Distance (FID)~\cite{heusel2017GANs} rely. 
Since nominally trained DNNs are generally vulnerable to adversarial attacks~\cite{croce2020reliable}, it is expected that DNN-based metrics for GANs also inherit these vulnerabilities.
Second, current adversarial attacks proposed in the literature are mainly designed at the instance level (\eg fooling a DNN into misclassifying a particular instance), while GAN metrics are distribution-based. 
Therefore, attacking these distribution-based metrics requires extending attack formulations from the paradigm of instances to that of distributions.

In this paper, we analyze the robustness of GAN metrics and recommend solutions to improve their robustness. 
We first attempt to assess the robustness of the quality measures used to evaluate GANs. 
We check whether such metrics are actually measuring the quality of image distributions by testing their vulnerability against additive pixel perturbations. 
While these metrics aim at measuring perceptual quality, we find that they are extremely brittle against imperceptible but carefully-crafted perturbations. 
We then assess the judgment of such metrics on the image distributions generated by StyleGANv2~\cite{Karras_2020_CVPR} when its input is subjected to perturbations. 
While the output of GANs is generally well behaved, we still observe that such metrics provide inconsistent judgments where, for example, FID favors an image distribution with significant artifacts over more naturally-looking distributions.
At last, we endeavor to reduce these metrics' vulnerability by incorporating robustly-trained models. 

We summarize our contributions as follows:
\begin{itemize}
    \item We are the first to provide an extensive experimental evaluation of the robustness of the Inception Score (IS) and the Fréchet Inception Distance~(FID) against additive pixel perturbations. 
    We propose two instance-based adversarial attacks that generate distributions of images that fool both IS and FID. 
    For example, we show that perturbations $\delta$ with a small budget (\ie $\|\delta\|_\infty \leq  0.01$) are sufficient to increase the FID between ImageNet~\cite{deng2009imagenet} and a perturbed version of ImageNet to $\sim 7900$, while also being able to generate a distribution of random noise images whose FID to ImageNet is $1.05$.
    We illustrate both cases in Figure~\ref{fig:pull}.
 
    \item We extend our evaluation to study the sensitivity of FID against perturbations in the latent space of state-of-the-art generative models. 
    In this setup, we show the vulnerability of both StyleGANv2 and FID against perturbations in both its $z$- and $w$- spaces. 
    We found that FID provides inconsistent evaluation of the distribution of generated images compared to their visual quality. 
    Moreover, our attack in the latent space causes StyleGANv2 to generate images with significant artifacts, showcasing the vulnerability of StyleGANv2 to additive perturbations in the latent space.
    
    \item We propose to improve the reliability of FID by using adversarially-trained models in its computation. 
    Specifically, we replace the traditional Inception model with its adversarially-trained counterpart to generate the embeddings on which the FID is computed. 
    We show that our robust metric, dubbed R-FID, is more resistant against pixel perturbations than the regular FID.
    
    \item Finally, we study the properties of R-FID when evaluating different GANs. 
    We show that R-FID is better than FID at distinguishing generated fake distributions from real ones.
    Moreover, R-FID provides more consistent evaluation under perturbations in the latent space of StyleGANv2. 
\end{itemize}

\section{Related Work}
\vspace{5pt}\noindent\textit{GANs and Automated Assessment.} GANs~\cite{goodfellow2014generative} have shown remarkable generative capabilities, specially in the domain of images~\cite{Karras_2019_CVPR,Karras_2020_CVPR,brock2018large}.
Since the advent of GANs, evaluating their generative capabilities has been challenging~\cite{goodfellow2014generative}.
This challenge spurred research efforts into developing automated quantitative measures for GAN outputs.
Metrics of particular importance for this purpose are the Inception Score (IS), introduced in~\cite{salimans2016improved}, and the Fréchet Inception Distance~(FID), introduced in~\cite{heusel2017GANs}.
Both metrics leverage the ImageNet-pretrained Inception architecture~\cite{szegedy2016rethinking} as a rough proxy for human perception.
The IS evaluates the generated images by computing conditional class distributions with Inception and measuring (1) each distribution's entropy---related to Inception's certainty of the image content, and (2) the marginal's entropy---related to diversity across generated images.
Noting the IS does not compare the generated distribution to the (real world) target distribution, Heusel~\etal~\cite{heusel2017GANs} proposed the FID.
The FID compares the generated and target distributions by (1) assuming the Inception features follow a Gaussian distribution and (2) using each distribution's first two moments to compute the Fréchet distance.
Further, the FID was shown to be more consistent with human judgement~\cite{shmelkov2018good}.

Both the original works and later research criticized these quantitative assessments.
On one hand, IS is sensitive to weight values, noisy estimation when splitting data, distribution shift from ImageNet, susceptibility to adversarial examples, image resolution, difficulty in discriminating GAN performance, and vulnerability to overfitting~\cite{barratt2018note,salimans2016improved,borji2019pros,xu2018empirical}.
On the other hand, FID has been criticized for its over-simplistic assumptions (``Gaussianity'' and its associated two-moment description), difficulty in discriminating GAN performance, and its inability to detect overfitting~\cite{borji2019pros,lucic2018gans,xu2018empirical}.
Moreover, both IS and FID were shown to be biased to both the number of samples used and the model to be evaluated \cite{chong2020effectively}.
In this work, we provide extensive empirical evidence showing that both IS and FID are not robust against perturbations that modify image quality.
Furthermore, we also propose a new \textit{robust} FID metric that enjoys superior robustness.

\vspace{5pt}\noindent\textit{Adversarial Robustness.} While DNNs became the \textit{de facto} standard for image recognition, researchers found that such DNNs respond unexpectedly to small changes in their input~\cite{szegedy2014intriguing,goodfellow2015explaining}.
In particular, various works~\cite{carlini2017towards,madry2018towards} observed a widespread vulnerability of DNN models against input perturbations that did not modify image semantics.
This observation spurred a line of research on adversarial attacks, aiming to develop procedures for finding input perturbations that fool DNNs~\cite{croce2020reliable}.
This line of work found that these vulnerabilities are pervasive, casting doubt on the nature of the impressive performances of DNNs.
Further research showed that training DNNs to be robust against these attacks~\cite{madry2018towards} facilitated the learning of perceptually-correlated features~\cite{ilyas2019adversarial,engstrom2020adversarial}.
Interestingly, a later work~\cite{santurkar2019image} even showed that such learnt features could be harnessed for image synthesis tasks.
In this work, we show (1) that DNN-based scores for GANs are vulnerable against adversarial attacks, and (2) how these scores can be ``robustified'' by replacing nominally trained DNNs with robustly trained ones.
\section{Robustness of IS and FID}\label{sec:robustness-of-fid}
To compare the output of generative models, two popular metrics are used: the \emph{Inception Score} (IS) and the \emph{Fréchet Inception Distance} (FID). 
These metrics depend only on the statistics of the distribution of generated images in an ImageNet-pretrained Inception's embedding space, raising the question:
\begin{center}
    \emph{What do quality measures for generative models, such as IS and FID, tell us about image quality?}
\end{center}
We investigate this question from the robustness perspective.
In particular, we analyze the sensitivity of these metrics to carefully crafted perturbations.
We start with preliminary background about both metrics.

\subsection{Preliminaries}
We consider the standard image generation setup where a generator $G: \mathbb R^{d_z} \rightarrow \mathbb R^{d_x}$ receives a latent code $z \in  \mathbb R^{d_z}$ and outputs an image $x \in \mathbb R^{d_x}$.
Upon training, $G$ is evaluated based on the quality of the generated distribution of images $\mathcal D_G$ by computing either the IS~\cite{salimans2016improved} or the FID~\cite{heusel2017GANs}.
Both metrics leverage an ImageNet-pretrained~\cite{deng2009imagenet} InceptionV3 \cite{szegedy2016rethinking}.
Salimans~\etal~\cite{salimans2016improved} proposed measuring the perceptual quality of the generated distribution $\mathcal D_G$ by computing the IS as:
\begin{equation}\label{eq: IS-formula}
    \text{IS}(\mathcal D_G) = \exp\left(\:\mathbb E_{x\sim \mathcal D_G}\left(\text{KL}\left(p(y|x)\:||\:p(y)\right)\right)\:\right),
\end{equation}
where $p(y|x)$ is the output probability distribution of the pretrained Inception model.
While several works have argued about the effectiveness of the IS and its widely-used implementation \cite{barratt2018note}, its main drawback is that it disregards the relation between the generated distribution, $\mathcal D_G$, and the real one, $\mathcal D_R$, used for training $G$ \cite{heusel2017GANs}.
Consequently, Heusel \etal proposed the popular FID, which involves the statistics of the real distribution.
In particular, FID assumes that the Inception features of an image distribution $\mathcal D$ follow a Gaussian distribution with mean $\mu_\mathcal D$ and covariance $\Sigma_\mathcal D$, and it measures the squared Wasserstein distance between the two Gaussian distributions of real and generated images.
Hence, $\text{FID}(\mathcal D_R, \mathcal D_G)$, or $\text{FID}$ for short, can be calculated as:
\begin{equation}\label{eq: FID-formula}
    \text{FID} = \| \mu_R - \mu_G\|^2 + \text{Tr}\left(\Sigma_R + \Sigma_G - 2(\Sigma_R  \Sigma_G)^{\nicefrac{1}{2}}\right),
\end{equation}
where $._R, ._G$ are the statistics of the real and generated image distributions, respectively, and $\text{Tr}(\cdot)$ is the trace operator. 
Note that the statistics of both distributions are empirically estimated from their corresponding image samples. 
In principle, FID measures how close (realistic) the generated distribution $\mathcal D_G$ is to $\mathcal D_R$.
We remark that the FID is the \emph{de facto} metric for evaluating image generation-related tasks.
Therefore, our study focuses mostly on FID.

We note here that both the IS and the FID are oblivious to $G$'s training process and can be computed to compare two arbitrary sets of images $\mathcal D_R$ and $\mathcal D_G$.
In generative modeling, this is typically a set of real images (photographs) and a set of generated images.
However, it is also possible to compare two sets of photographs, two sets of generated images, manipulated photographs with real photographs, \emph{etc}.
This flexibility allows us to study these metrics in a broader context next, where no generative model is involved.

\subsection{Robustness under Pixel Perturbations}\label{subsec: robustness under pixel perturbations}

We first address the question presented earlier in Section~\ref{sec:robustness-of-fid} by analyzing the sensitivity of IS and FID to additive pixel perturbations. 
In particular, we assume $\mathcal D_R$ to be either CIFAR10~\cite{krizhevsky2009learning} or ImageNet~\cite{deng2009imagenet} and ask: \textbf{(i)} can we generate a distribution of imperceptible additive perturbations $\delta$ that deteriorates the scores for $\mathcal D_G = \mathcal D_R + \delta$? 
Or, alternatively, \textbf{(ii)} can we generate a distribution of low visual quality images, \ie noise images, that attain good quality scores? 
If the answer is yes to both questions, then FID and IS have limited capacity for providing information about image quality in the worst case.

\subsubsection{Good Images - Bad Scores}\label{sec:goodims_badscores}
\begin{figure}[t]
    \centering
    \includegraphics[width=0.65\columnwidth]{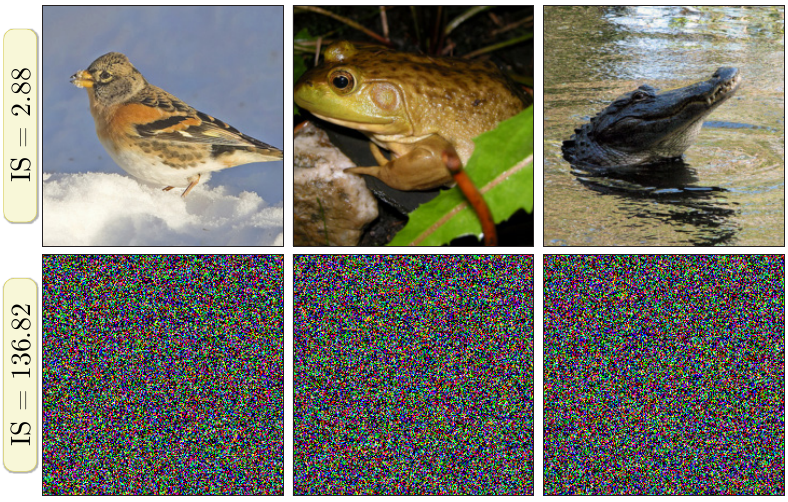}
    \caption{
    \textbf{Sensitivity of Inception Score (IS) against pixel perturbations.} 
    \textit{First row:} real-looking images (sampled from $\mathcal{D}_G = \mathcal D_R + \delta$) with a low IS (below 3). 
    \textit{Second row:} random noise images with a high IS (over 135).}
    \label{fig:is_pixel}
\end{figure}
We aim at constructing a distribution of real-looking images with \emph{bad} quality measures, \ie low IS or high FID. 
While both metrics are distribution-based, we design instance-wise proxy optimization problems to achieve our goal. 

\noindent\textit{Minimizing IS.} 
Based on Eq. \eqref{eq: IS-formula}, one could minimize the IS by having both the posterior $p(y|x)$ and the prior $p(y)$ be the same distribution. 
Assuming that $p(y)$ is a uniform distribution, we minimize the IS by maximizing the entropy of $p(y|x)$.
Therefore, we can optimize a perturbation $\delta^*$ for each real image $x_r\sim \mathcal D_R$ by solving the following problem:
\begin{equation}\label{eq:minimize-IS}
    \begin{aligned}
      &\delta^* = \argmax_{\|\delta\|_\infty \leq \epsilon} ~\mathcal L_{\text{ce}}\left(p(y|x_r+\delta), \hat{y}\right), \\
      &\text{s.t. } \hat y = \argmax_i~ p^i(y|x_r + \delta),
    \end{aligned}
\end{equation}
where $\mathcal L_{\text{ce}}$ is the Cross Entropy loss.
We solve the problem in Eq.~\eqref{eq:minimize-IS} with 100 steps of Projected Gradient Descent (PGD) and zero initialization.
We then compile the distribution $\mathcal D_G$, where each image $x_g = x_r + \delta^*$ is a perturbed version of the real dataset $\mathcal D_R$. 
Note that our objective aims to minimize the network's confidence in predicting all labels for each $x_g$. In doing so, both $p(y|x_g)$ and $p(y)$ tend to converge  to a uniform distribution, thus,  minimizing the KL divergence between them and effectively lowering the IS.
Note how $\epsilon$ controls the allowed perturbation amount for each image $x_r$. 
Therefore, for small $\epsilon$ values, samples from  $\mathcal D_G$ and $\mathcal D_R$ are perceptually indistinguishable.

\begin{table}[t]
\centering
\caption{\textbf{Robustness of IS and FID against pixel perturbations.} 
We assess the robustness of IS and FID against perturbations with a limited budget $\epsilon$ on CIFAR10 and ImageNet. 
In the last row, we report the IS and FID of images with carefully-designed random noise having a resolution similar to CIFAR10 and ImageNet.}
\centering
\begin{tabular}{x{2cm}|x{1.2cm}x{1.2cm}|x{1.2cm}x{1.2cm}}
\toprule 
\midrule
 \multirow{2}{*}{$\epsilon$}& \multicolumn{2}{c|}{CIFAR10} & \multicolumn{2}{c}{ImageNet} \\
 & IS & FID & IS & FID \\
 \midrule
0.00 & 11.54 & 0.00 & 250.74 & 0.00 \\
 $5\times10^{-3}$  &2.62 &142.45 &3.08 &3013.33 \\
 $0.01$  & 2.50 &473.19 &2.88 &7929.01 \\
\midrule
random noise &94.87 &9.94 &136.82 & 1.05\\
\midrule
\bottomrule
\end{tabular}\label{tb:FID-pixel}
\vspace{-0.5cm}
\end{table}

\noindent\textit{Maximizing FID.} 
Next, we extend our attack setup to the more challenging FID. 
Given an image $x$, we define $f(x): \mathbb R^{d_x} \rightarrow \mathbb R^{d_e}$ to be the output embedding of an Inception model. 
We aim to maximize the FID by generating a perturbation $\delta$ that pushes the embedding of a real image away from its original position. 
In particular, for each $x_r \sim \mathcal D_R$, we aim to construct $x_g = x_r + \delta^*$ where:
\begin{equation}\label{eq:maximize-fid}
    \delta^* = \argmax_{ \| \delta \|_\infty \leq \epsilon} ~\left\| f(x_r) - f(x_r +  \delta) \right\|_2 .
\end{equation}
In our experiments, we solve the optimization problem in Eq. \eqref{eq:maximize-fid} with 100 PGD steps 
and a randomly initialized $\delta$ \cite{madry2018towards}. 
Maximizing this objective indirectly maximizes FID's first term (Eq.~\eqref{eq: FID-formula}), while resulting in a distribution of images $\mathcal D_G$ that is visually indistinguishable from the real $\mathcal D_R$ for small $\epsilon$ values. 

\noindent\textit{Experiments.} 
We report our results in Table~\ref{tb:FID-pixel}. 
Our simple yet effective procedure illustrates how both metrics are very susceptible to attacks. 
In particular, solving the problem in Eq.~\eqref{eq:minimize-IS} yields a distribution of noise that significantly decreases the IS from 11.5 to 2.5 in CIFAR10 and from 250.7 to 2.9 in ImageNet. 
We show a sample from $\mathcal D_G$ in Figure~\ref{fig:is_pixel}, first row. 
Similarly, our optimization problem in Eq.~\eqref{eq:maximize-fid} can create imperceptible perturbations that maximize the FID to $\approx$7900 between ImageNet and its perturbed version (examples shown in Figure~\ref{fig:pull}).

\subsubsection{Bad Images - Good Scores}\label{subsec:bad-images-good-quality}
While the previous experiments illustrate the vulnerability of both the IS and FID against small perturbations (\ie good images with bad scores), here we evaluate if the converse is also possible, \ie bad images with good scores. 
In particular, we aim to construct a distribution of noise images (\eg second row of Figure~\ref{fig:is_pixel}) that enjoys good scores (high IS or low FID).

\noindent\textit{Maximizing IS.} 
The IS has two terms: Inception's confidence on classifying a generated image, \ie $p(y|x_g)$, and the diversity of the generated distribution of predicted labels, \ie $p(y)$. 
One can maximize the IS by generating a distribution $\mathcal D_G$ such that: \textbf{(i)} each $x_g \sim \mathcal D_G$ is predicted with high confidence, and \textbf{(ii)} the distribution of predicted labels is uniform across Inception's output $\mathcal Y$. 
To that end, we propose the following procedure for constructing such $\mathcal D_G$. 
For each $x_g$, we sample a label $\hat y \sim \mathcal Y$ uniformly at random and solve the problem:
\begin{equation}\label{eq: maximize IS}
x_g = \argmin_x ~\mathcal{L}_{ce}(p(y|x), \hat y).
\end{equation}
In our experiments, we solve the problem in Eq.~\eqref{eq: maximize IS} with 100 gradient descent steps and random initialization for $x$.

\noindent\textit{Minimizing FID.} 
Here, we analyze the robustness of FID against such a threat model. We follow a similar strategy to the objective in Eq.~\eqref{eq:maximize-fid}. 
For each image $x_r \sim \mathcal D_R$, we intend to construct $x_g$ such that:
\begin{equation} \label{eq: minimizing-fid}
    x_g = \argmin_x ~\left\|f(x) - f(x_r)\right\|_2
\end{equation}
with a randomly initialized $x$. 
In our experiments, we solve Eq. \eqref{eq: minimizing-fid} with 100 gradient descent steps.
As such, each $x_g$ will have a similar Inception representation to a real-world image, \ie $f(x_g) \approx f(x_r)$, while being random noise. 

\paragraph{Experiments.} 
We report our results in the last row of Table \ref{tb:FID-pixel}. 
Both the objectives in Eqs.~\eqref{eq: maximize IS} and \eqref{eq: minimizing-fid} are able to fool the IS and FID, respectively. 
In particular, we are able to generate distributions of noise images with resolutions $32\times32$ and $224\times224$ (\ie CIFAR10 and ImageNet resolutions) but with IS of 94 and 136, respectively. 
We show a few qualitative samples in the second row of Figure~\ref{fig:is_pixel}. 
Furthermore, we generate noise images that have embedding representations very similar to those of CIFAR10 and ImageNet images. This lowers the FID of both datasets to 9.94 and 1.05, respectively (examples are shown in Figure \ref{fig:pull}).


\subsection{Robustness under Latent Perturbations}\label{subsec: robustness under latent perturbations}
In the previous section, we established the vulnerability of both the IS and FID against pixel perturbations. 
Next, we investigate the vulnerability against perturbations in a GAN's latent space. 
Designing such an attack is more challenging in this case, since images can only be manipulated indirectly, and so there are fewer degrees of freedom for manipulating an image. 
To that end, we choose $G$ to be the state of the art generator StyleGANv2~\cite{Karras_2019_CVPR} trained on the standard FFHQ dataset~\cite{Karras_2019_CVPR}. 
We limit the investigation to the FID metric, as IS is not commonly used in the context of unconditional generators, such as StyleGAN. 
Note that we always generate $70$k samples from $G$ to compute the FID.

Recall that our generator $G$ accepts a random latent vector $z \sim \mathcal{N}(0, \text{I})$\footnote{The appendix presents results showing that sampling $z$ from different distributions still yields good looking StyleGANv2-generated images.} and maps it to the more expressive latent space $w$, which is then fed to the remaining layers of $G$. 
It is worthwhile to mention that ``truncating'' the latent $w$ with a pre-computed $\bar{w}$\footnote{$\bar{w}$ is referred to as the mean of the $w$-space. It is computed by sampling several latents $z$ and averaging their representations in the $w$-space.} and constant $\alpha\in \mathbb R$ (\ie replacing $w$ with $\alpha w + (1-\alpha)\bar{w}$) controls both the quality and diversity of the generated images~\cite{Karras_2019_CVPR}.

\begin{figure*}[t]
    \centering
    \includegraphics[width=\textwidth]{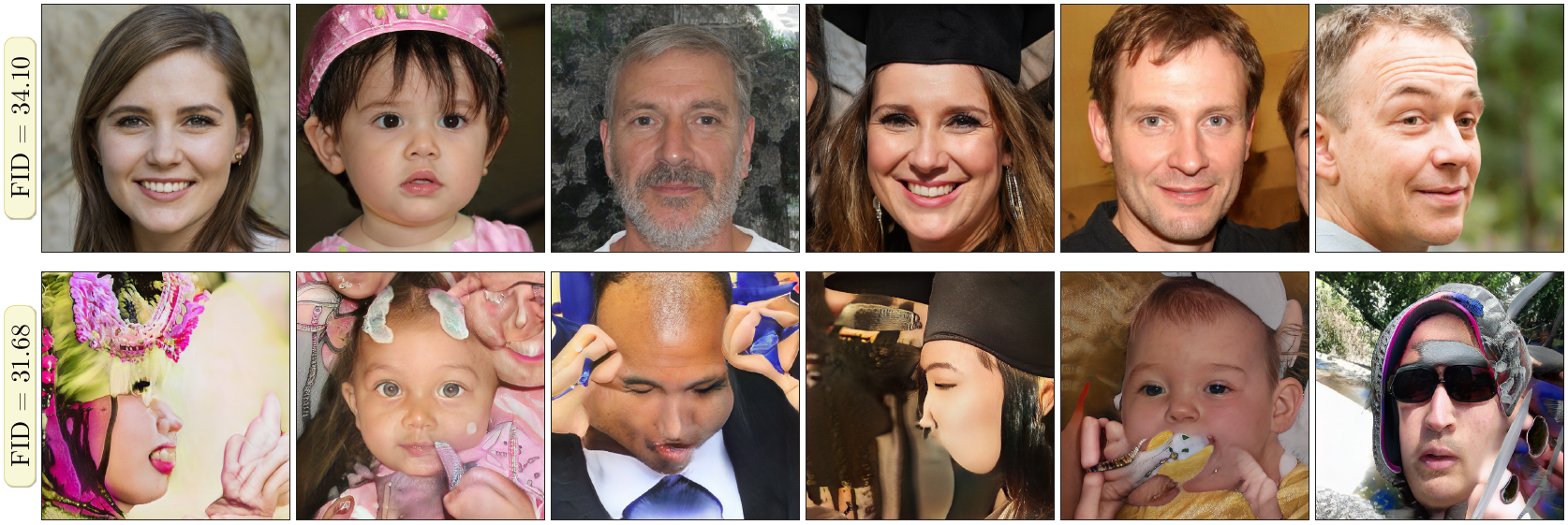}
    \caption{
    \textbf{Effect of attacking truncated StyleGANv2's latent space on the Fréchet Inception Distance (FID).}
    We conduct attacks on the latent space of StyleGANv2 and record the effect on the FID.
    We display the resulting samples of these attacks for two truncation values, $\alpha=0.7$ (top row) and $\alpha=1.0$ (bottom row).
    Despite the stark differences in realism between the images in the top and bottom rows---\ie the top row's remarkable quality and the bottom row's artifacts---the FID to FFHQ reverses this ranking, wherein the bottom row is judged as \textit{farther} away from FFHQ than the top row.
    }
    \label{fig:fid_latent}
\end{figure*}

\paragraph{Effect of Truncation on FID.} 
We first assess the effect of the truncation level $\alpha$ on both image quality and FID. 
We set $\alpha \in [0.7, 1.0, 1.3]$ and find FIDs to be $[21.81, 2.65, 9.31]$, respectively. 
Based on our results, we assert the following observation: while the visual quality of generated images at higher truncation levels, \eg $\alpha = 0.7$, is better and has fewer artifacts than the other $\alpha$ values, the FID does not reflect this fact, showing lower (better) values for $\alpha\in\{1.0, 1.3\}$. 
We elaborate on this observation with qualitative experiments in the appendix.

\paragraph{FID-Guided Sampling.} 
Next, we extend the optimization problem in Eq.~\eqref{eq:maximize-fid} from image to latent perturbations. 
In particular, we aim at constructing a perturbation $\delta^*_z$ for each sampled latent $z$ by solving:
\begin{equation}\label{eq:maximize-fid-latent}
\begin{aligned}
  \delta^*_z &= \argmax_\delta ~\left\|f( G(z+\delta) ) - f(x_r) \right\|_2.
\end{aligned}
\end{equation}
Thus, $\delta^*_z$ perturbs $z$ such that $G$ produces an image whose embedding differs from that of real image $x_r$. 
We solve the problem in Eq.~\eqref{eq:maximize-fid-latent} for $\alpha \in \{0.7, 1.0\}$.

\paragraph{Experiments.} 
We visualize our results in Figure \ref{fig:fid_latent} accompanied with their corresponding FID values (first and second rows correspond to $\alpha=0.7 \text{ and } 1.0$, respectively). 
While our attack in the latent space is indeed able to significantly increase the FID (from 2.65 to 31.68 for $\alpha=1.0$ and 21.33 to 34.10 for $\alpha=0.7$), we inspect the results and draw the following conclusions. 
\textbf{(i)} FID provides inconsistent evaluation of the generated distribution of images. 
For example, while both rows in Figure \ref{fig:fid_latent} have comparable FID values, the visual quality is significantly different. 
This provides practical evidence of this metric's unreliability in measuring the performance of generative models. 
\textbf{(ii)} Adding crafted perturbations to the input of a state of the art GAN deteriorates the visual quality of its output space (second row in Figure \ref{fig:fid_latent}). 
This means that GANs are also vulnerable to adversarial attacks. 
This is confirmed in the literature for other generative models such as GLOW~\cite{NEURIPS2018_d139db6a,pmlr-v108-pope20a}. 
Moreover, we can formulate a problem similar to Eq.~\eqref{eq:maximize-fid-latent} but with the goal of perturbing the $w$-space instead of the $z$-space.
We leave results of solving this formulation for different $\alpha$ values to the appendix.

\paragraph{\textbf{Section Summary.}} 
In this section, we presented an extensive experimental evaluation investigating if the quality measures (IS and FID) of generative models actually measure the perceptual quality of the output distributions. 
We found that such metrics are extremely vulnerable to pixel perturbations. 
We were able to construct images with very good scores but no visual content (Section~\ref{subsec:bad-images-good-quality}), as well as images with realistic visual content but very bad scores (Section~\ref{sec:goodims_badscores}). 
We further studied the sensitivity of FID against perturbations in the latent space of StyleGANv2 (Section~\ref{subsec: robustness under latent perturbations}), allowing us to establish the inconsistency of FID under this setup as well.
Therefore, we argue that such metrics, while measuring useful properties of the generated distribution, lead to questionable assessments of the visual quality of the generated images.

\section{R-FID: Robustifying the FID}
After establishing the vulnerability of IS and FID to perturbations, we analyze the cause of such behavior and propose a solution. 
We note that, while different metrics have different formulations, they rely on a pretrained Inception model that could potentially be a leading cause of such vulnerability.
This observation suggests the following question:
\begin{center}
    \emph{Can we robustify the FID by replacing its Inception\\component with a robustly trained counterpart?}
\end{center}
We first give a brief overview of adversarial training.

\subsection{Leveraging Adversarially Trained Models}
Adversarial training is arguably the \emph{de facto} procedure for training robust models against adversarial attacks. 
Given input-label pairs $(x, y)$ sampled from a training set $\mathcal D_{tr}$, $\ell_2$-adversarial training solves the following min-max problem:
\begin{equation}\label{eq:adv-training}
\min_{\theta}~ \mathbb E_{(x, y) \sim \mathcal D_{tr}}\left[ \max_{\|\delta\|_2 \leq \kappa} \mathcal L \left(x+\delta, y; \theta\right) \right] 
\end{equation}
\noindent for a given loss function $\mathcal L$ to train a robust network with parameters $\theta$. 
We note that $\kappa$ controls the robustness-accuracy trade-off: models trained with larger $\kappa$ tend to have higher robust accuracy (accuracy under adversarial attacks) and lower clean accuracy (accuracy on clean images). 
Since robust models are expected to resist pixel perturbations, we expect such models to inherit robustness characteristics against the attacks constructed in Section \ref{subsec: robustness under pixel perturbations}. 
Moreover, earlier works showed that robustly-trained models tend to learn more semantically-aligned and invertible features~\cite{ilyas2019adversarial}. 
Therefore, we hypothesize that replacing the pretrained Inception model with its robustly trained counterpart could increase FID's sensitivity to the visual quality of the generated distribution (\ie robust against attacks in Section~\ref{subsec: robustness under latent perturbations}).

To that end, we propose the following modification to the FID computation. 
We replace the pretrained Inception model with a robustly trained version on ImageNet following Eq.~\eqref{eq:adv-training} with $\kappa \in \{64, 128\}$. 
The training details are left to the appendix. 
We refer to this alternative as R-FID, and analyze its robustness against perturbations next. 

\begin{figure}[t]
    \centering
    \includegraphics[width=0.7\columnwidth]{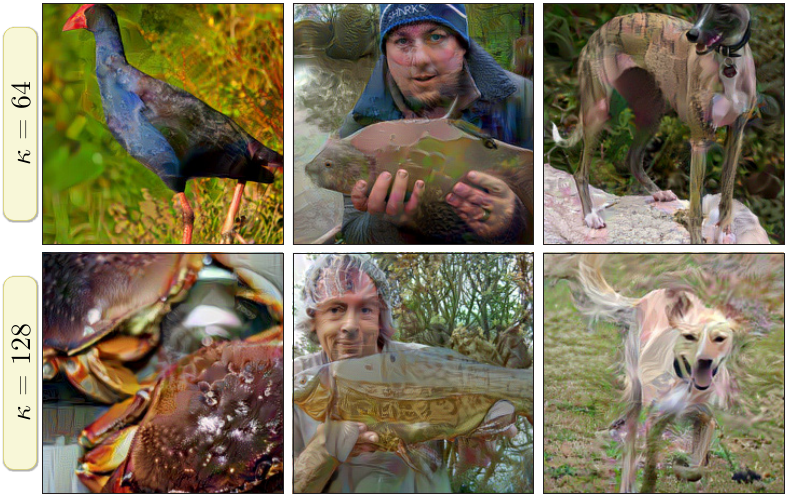}
    \caption{
    \textbf{Attacking R-FID with pixel perturbations.}
    We attack two variants of R-FID ($\kappa = 64$ and $\kappa = 128$) and visualize samples from the resulting datasets.
    Attempting to fool these R-FIDs at the pixel level yields perturbations that correlate with semantic patterns, in contrast to those obtained when attempting to fool the standard FID (as shown in Figure~\ref{fig:pull}).
    }
    \label{fig:rfid_pixel}
\end{figure}

\begin{table}[t]
\centering
\caption{\textbf{R-FID against attacks in the pixel space.} We study the robustness of R-FID against the adversarial attacks in Eq. \ref{eq:maximize-fid}. }
\centering
\begin{tabular}{x{0.8cm}|x{1.3cm}x{1.3cm}|x{1.3cm}x{1.3cm}}
\toprule 
\midrule
 \multirow{2}{*}{$\epsilon$}& \multicolumn{2}{c|}{CIFAR10} & \multicolumn{2}{c}{ImageNet} \\
     & $\kappa=64$ & $\kappa=128$ & $\kappa=64$ & $\kappa=128$ \\
\midrule

 $0.01$  &1.5  &0.3 &21.0 &4.5 \\
 $0.02$  &20.7 &7.8 &293.8   &92.1 \\
 $0.03$  &46.4 &19.7 &657.9  &264.6 \\
\midrule
\bottomrule
\end{tabular}\label{tb:RFID-pixel}
\end{table}



\subsection{R-FID against Pixel Perturbations}\label{subsec:rfid-pixel}
We first test the sensitivity of R-FID against additive pixel perturbations. 
For that purpose, we replace the Inception with a robust Inception, and repeat the experiments from Section~\ref{sec:goodims_badscores} to construct real images with bad scores. 
We conduct experiments on CIFAR10 and ImageNet with $\epsilon \in \{0.01, 0.02, 0.03\}$ for the optimization problem in Eq.~\eqref{eq:maximize-fid}, and we report the results in Table \ref{tb:RFID-pixel}. 
We observe that the use of a robustly-trained Inception significantly improves robustness against pixel perturbations. 
Our robustness improvement for the same value of $\epsilon=0.01$ is of 3 orders of magnitude (an FID of 4 for $\kappa=128$ compared to 7900 reported in Table~\ref{tb:FID-pixel}). 
While both models consistently provide a notable increase in robustness against pixel perturbations, we find that the model most robust to adversarial attacks (\ie $\kappa=128$) is also the most robust to FID attacks. 
It is worthwhile to mention that this kind of robustness is expected since our models are trained not to alter their prediction under additive input perturbations. 
Hence, their feature space should enjoy robustness properties, as measured by our experiments. 
In Figure~\ref{fig:rfid_pixel} we visualize a sample from the adversarial distribution $\mathcal D_G$ (with $\epsilon=0.08$) when $\mathcal D_R$ is ImageNet. 
We observe that our adversaries while aiming only at pushing the feature representation of samples of $\mathcal D_G$ away from those of $\mathcal D_R$, are also more correlated with human perception. 
This finding aligns with previous observations in the literature, which find robustly-trained models have a more interpretable (more semantically meaningful) feature space~\cite{ilyas2019adversarial,engstrom2020adversarial}. 
We leave the evaluation under larger values of~$\epsilon$, along with experiments on unbounded perturbations, to the appendix. 

\begin{table}[t]
\centering
\caption{\textbf{Truncation's effect on R-FID.} We study how truncation affects R-FID against FFHQ (first two rows), and across different truncation levels~(last two rows).}
\centering
\begin{tabular}{x{2.8cm}|x{1.4cm}x{1.4cm}x{1.4cm}}
\toprule 
\midrule
$(\mathcal D_G(\alpha)$, $\mathcal D_R)$   & 0.7 & 0.9 & 1.0 \\
\midrule
$\kappa=64$     & 98.3 & 90.0 & 88.1  \\
$\kappa=128$     & 119.9 & 113.7 & 113.8  \\
\midrule
\midrule
$(\mathcal D_G(\alpha_i), \mathcal D_G(\alpha_j))$ & (0.7, 1.0)& (0.7, 0.9)& (0.9, 1.0) \\
\midrule
$\kappa=64$     & 10.5 & 4.9 & 0.48  \\
$\kappa=128$     & 9.9 & 4.6 & 0.46  \\
\midrule
\bottomrule
\end{tabular}\label{tb:RFID-ablation}
\end{table}

\subsection{R-FID under Latent Perturbations}\label{subsec:rfid-latent}
In Section \ref{subsec:rfid-pixel}, we tested R-FID's robustness against pixel-level perturbations. 
Next, we study R-FID for evaluating generative models. 
For this, we follow the setup in Section~\ref{subsec: robustness under latent perturbations} using an FFHQ-trained StyleGANv2 as generator $G$.  

\begin{figure*}[t]
    \centering
    \includegraphics[width=\textwidth]{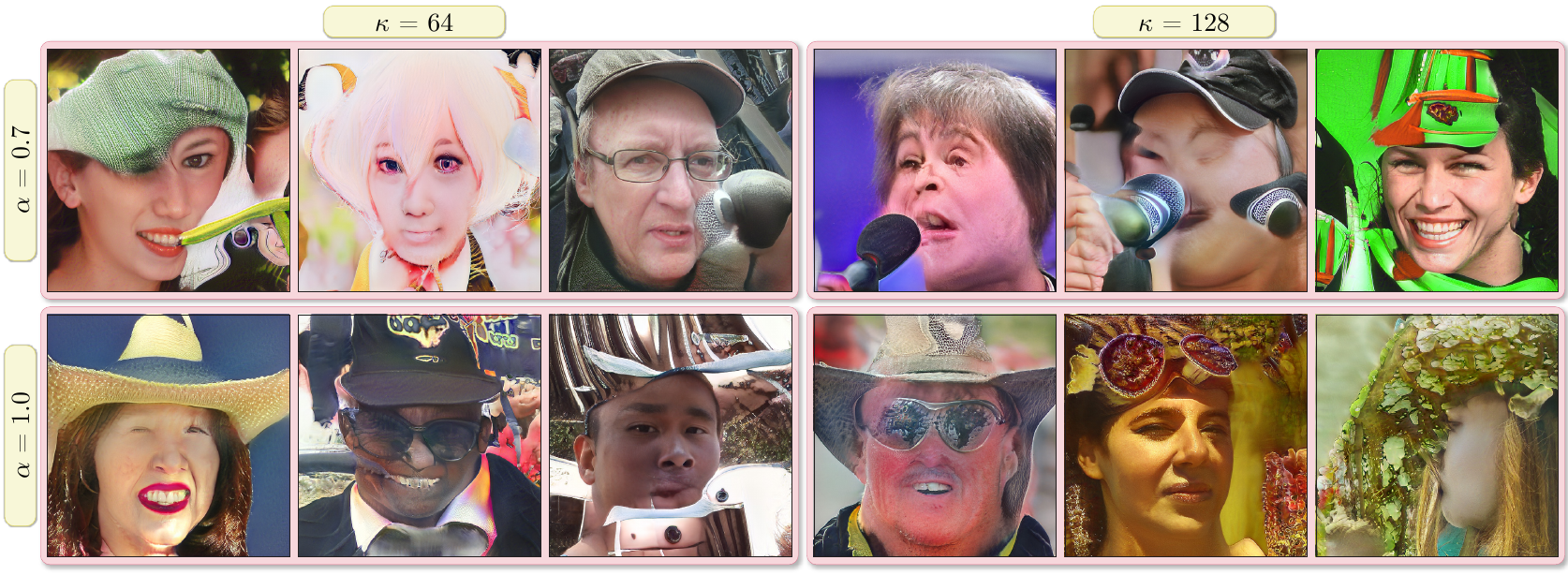}
    \caption{
    %
    \textbf{Robustness of R-FID against perturbations in StyleGANv2 latent space.}
    We conduct attacks on two variants of R-FID ($\kappa=64$ on the left, and $\kappa=128$ on the right) and two truncation values ($\alpha=0.7$ on the top, and $\alpha=1.0$ on the bottom) by perturbing the latent space. We also visualize samples from the generated distributions. 
    For the pairs $(\kappa, \alpha) \in \{(64, 0.7), (64, 1.0), (128, 0.7), (128, 1.0)\}$, we find corresponding R-FID values of \{128.1, 157.8, 126.6, 162,8\}.
    In contrast to the minimal changes required to fool the standard FID (Fig.~\ref{fig:fid_latent}), fooling the R-FID leads to a dramatic degradation in visual quality of the generated images.
    }
    \label{fig:rfid_latent}
\end{figure*}

\vspace{5pt}\noindent\textit{Effect of Truncation on R-FID.} 
Here, we analyze the R-FID when the generator is using different truncation levels. 
In particular, we choose $\alpha \in \{0.7, 0.9, 1.0\}$ and report  results in Table~\ref{tb:RFID-ablation}. 
We observe that the robust Inception model clearly distinguishes the distribution generated by StyleGANv2 from the FFHQ dataset, regardless of the truncation $\alpha$. 
In this case, we obtain an R-FID of 113.8, substantially larger than the 2.6 obtained when the nominally-trained Inception model is used. 
This result demonstrates that, while the visual quality of StyleGANv2's output is impressive, the generated image distribution is far from the FFHQ distribution. 
We further evaluate if the R-FID is generally large between any two distributions by measuring the R-FID between two distributions of images generated at two truncation levels $(\alpha_i, \alpha_j)$. 
Table \ref{tb:RFID-ablation} reports these results. 
We observe that \textbf{(i)} the R-FID between a distribution and itself is $\approx 0$, \eg  R-FID = $10^{-3}$ at (1.0, 1.0).
Please refer to the appendix for details.
\textbf{(ii)} The R-FID gradually increases as the image distributions differ, \eg R-FID at (0.9, 1.0) $<$ (0.7, 1.0). 
This observation validates that the large R-FID values found between FFHQ and various truncation levels are a result of the large separation in the embedding space that robust models induce between real and generated images.

\vspace{5pt}\noindent\textit{R-FID Guided Sampling.} 
Next, we assess the robustness of the R-FID against perturbations in the latent space of the generator $G$. 
For this purpose, we conduct the attack proposed in Eq.~\eqref{eq:maximize-fid-latent} with $f$ now being the robustly-trained Inception. 
We report results and visualize few samples in Figure~\ref{fig:rfid_latent}. 
We make the following observations. 
\textbf{(i)} While the R-FID indeed increases after the attack, the relative increment is far less than that of the non-robust FID. 
For example, R-FID increases by 44\% at $\kappa=64$ and $\alpha=0.7$ compared to an FID increase of 1000\% under the same setup. 
\textbf{(ii)} The increase in R-FID is associated with a significantly larger amount of artifacts introduced by the GAN in the generated images. 
This result further evidences the vulnerability of the generative model.
However, it also highlights the changes in the image distribution that are required to increase the R-FID. 
We leave the $w$- space formulation for the attack on the R-FID, along with its experiments, to the appendix.

\vspace{5pt}\noindent\textit{\textbf{Section Summary.}} 
In this section, we robustified the popular FID by replacing the pretrained Inception model with a robustly-trained version. 
We found this replacement results in a more robust metric (R-FID) against perturbations in both the pixel (Section~\ref{subsec:rfid-pixel}) and latent (Section~\ref{subsec:rfid-latent}) spaces. 
Moreover, we found that pixel-based attacks yield much more perceptually-correlated perturbations when compared to the attacks that used the standard FID (Figure~\ref{tb:RFID-pixel}). 
Finally, we observed that
changing R-FID values requires a more significant and notable distribution shift in the generated images (Figure~\ref{fig:rfid_latent}).

\subsection{R-FID against Quality Degradation}

\begin{table}[t]
\centering
\caption{\textbf{Sensitivity of R-FID against noise and blurring.} We measure R-FID $(\kappa=128)$ between ImageNet and a transformed version of it under Gaussian noise and blurring. 
As $\sigma$ increases, the image quality decreases and R-FID increases.}
\centering
\begin{tabular}{x{2.8cm}|x{1.1cm}x{1.1cm}x{1.1cm}x{1.1cm}}
\toprule 
\midrule
  $\nicefrac{\sigma_N}{\sigma_B}$ & \nicefrac{0.1}{1.0} & \nicefrac{0.2}{2.0} & \nicefrac{0.3}{3.0} & \nicefrac{0.4}{4.0}\\
 \midrule
Gaussian (N)oise & 16.65 & 61.33 & 128.8 & 198.3 \\
Gaussian (B)lur & 15.54 & 54.07 & 78.67 & 89.11\\
\midrule
\bottomrule
\end{tabular}\label{tb:RFID-quality-degradation}
\end{table}

At last, we analyze the effect of transformations that degrade image quality on R-FID. 
In particular, we apply Gaussian noise and Gaussian blurring on ImageNet and report the R-FID $(\kappa=128)$ between ImageNet and the degraded version in Table \ref{tb:RFID-quality-degradation}. 
Results show that as the quality of the images degrades~(\ie as $\sigma$ increases), the R-FID steadily increases. 
Thus, we find that R-FID is able to distinguish a distribution of images from its degraded version. 

\section{Discussion, Limitations, and Conclusions}
In this work, we demonstrate several failure modes of popular GAN metrics, specifically IS and FID. We also propose a robust counterpart of FID (R-FID), which mitigates some of the robustness problems and yields significantly more robust behavior under the same threat models.

Measuring the visual quality for image distributions has two components: (1) the statistical measurement (\eg Wasserstein distance) and (2) feature extraction using a pretrained model (\eg InceptionV3). 
A limitation of our work is that we only focus on the second part (the pretrained model).
As an interesting avenue for future work, we suggest a similar effort to assess the reliability of the statistical measurement as well, \ie analyzing and finding better and more robust alternatives to the Wasserstein distance.

Current metrics mainly focus on comparing the distribution of features. In these cases, visual quality is only hoped to be a side effect and not directly optimized for nor tested by these metrics. Developing a metric that
directly assesses visual quality remains an open problem that is not tackled by our work but is
recommended for future work.

\textbf{Acknowledgments.} This work was supported by the King Abdullah University of Science and Technology (KAUST) Office of Sponsored Research (OSR) under Award No. OSR-CRG2019-4033.

\clearpage
%
%
\bibliographystyle{splncs04}
\bibliography{egbib}
\newpage
\appendix

\section{Sampling $z$ Outside Standard Gaussian}
\begin{figure}
    \centering
    \includegraphics[width=\textwidth]{ 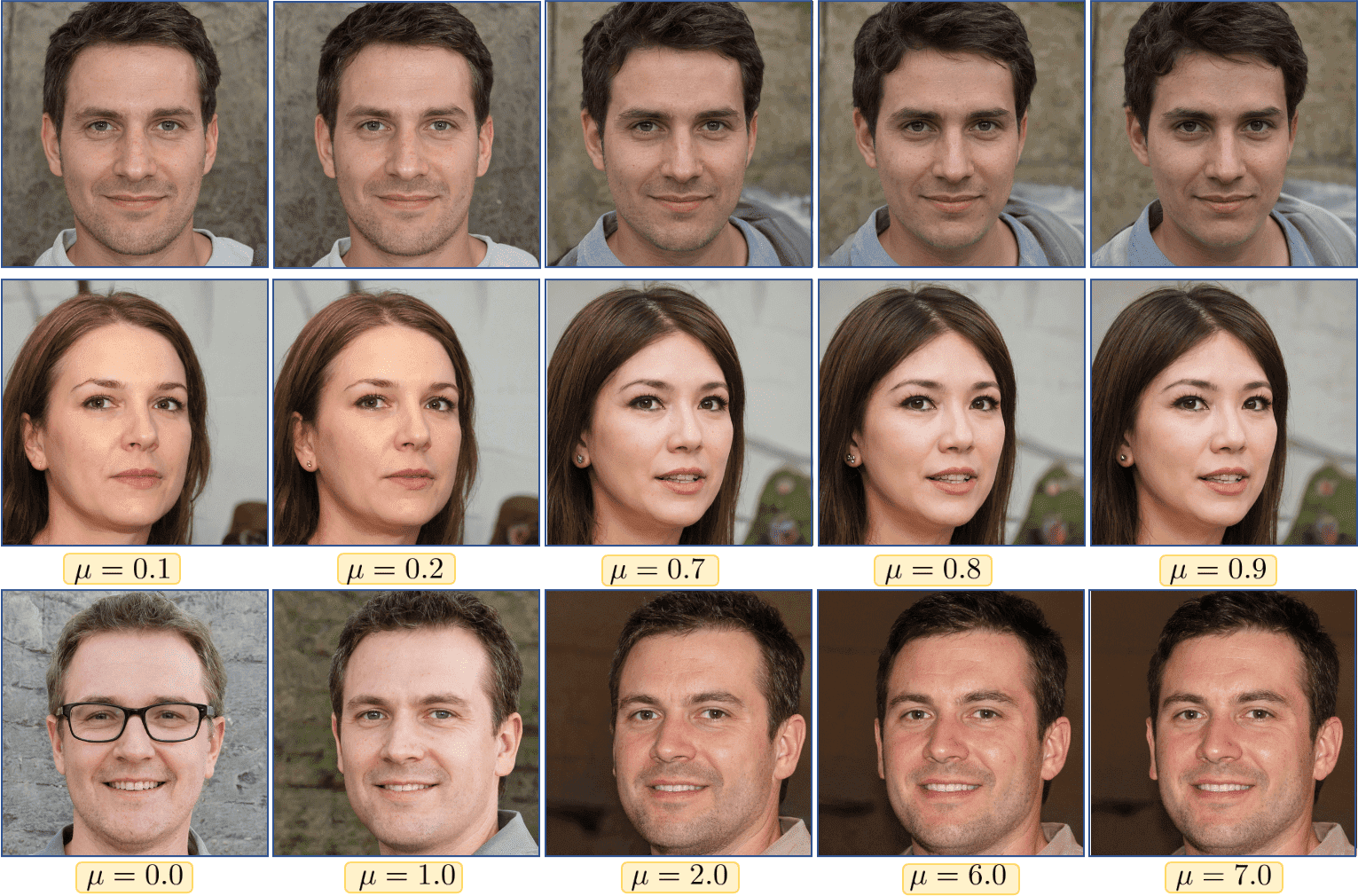}
    \caption{\textbf{Effect of shifting the mean of the Gaussian distribution on the output visual quality.} We notice that, for a truncation level $\alpha=0.5$, shifting the mean of the Gaussian distribution from which we sample the latent $z$ has a \emph{very minor} effect on the visual quality of the generated images.}
    \label{fig:shifting-mean}
\end{figure}

\begin{figure}
    \centering
    \includegraphics[width=\textwidth]{ 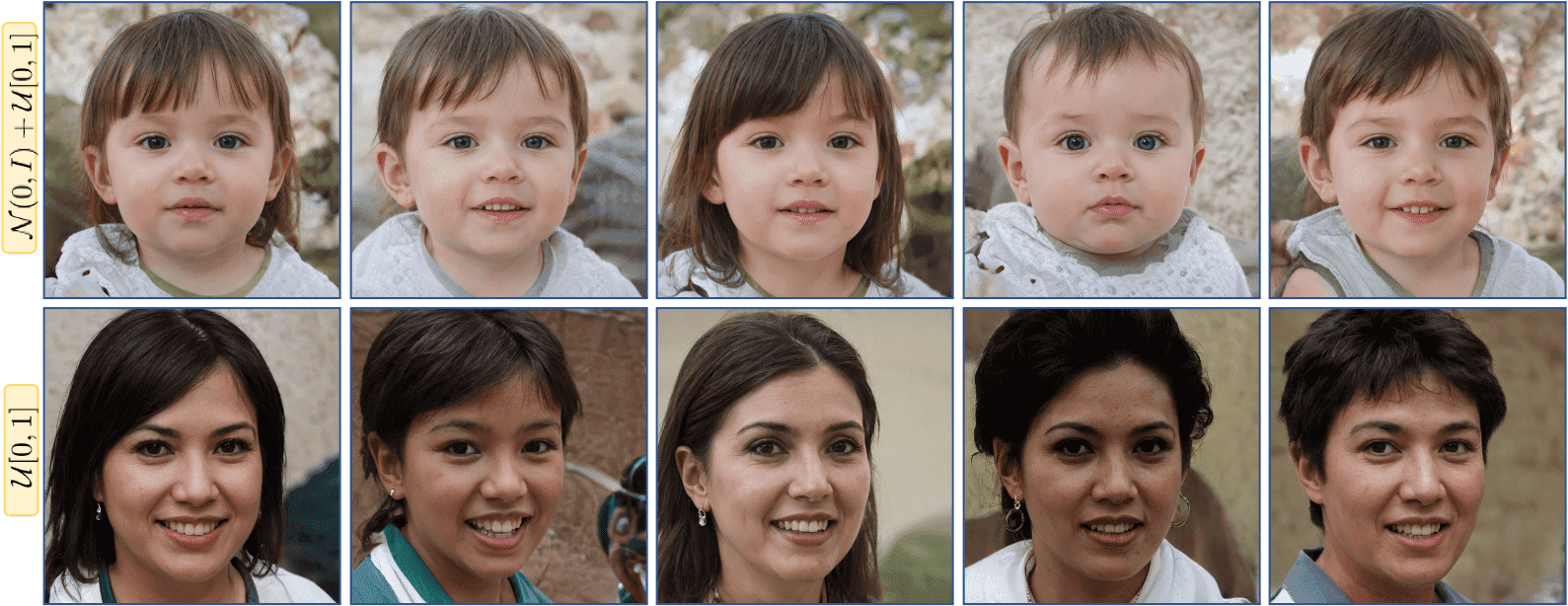}
    \caption{\textbf{Sampling from other distributions than standard Gaussian.} 
    We analyze the effect of adding a random uniform vector to the sampled $z$ from a standard Gaussian in the first row. 
    In the second row, we sample $z$ from a uniform distribution as opposed to the standard Gaussian. 
    In both cases, and for truncation level of $\alpha=0.5$, we note that StyleGANv2 is capable of producing output images with good visual quality.}
    \label{fig:uniform-sampling}
\end{figure}

In this section, we check the effect of sampling the latent $z$ from distributions other than the one used in training. In particular, instead of sampling $z$ from a standard Gaussian distribution, we try the following setups:
\begin{itemize}
    \item $z \sim \mathcal{N}(\mu, I)$ where $\mu \in \{0.1, 0.2, 0.7, 0.8, 0.9, 1.0, 2.0, 6.0, 7.0\}$.
    \item $z \sim \mathcal{N}(\mu, I) + \mathcal{U}[0, 1]$ where $\mathcal{U}$ is a uniform distribution.
    \item $z \sim \mathcal{U}[0, 1]$
\end{itemize}
We report the results in Figures \ref{fig:shifting-mean} and \ref{fig:uniform-sampling}, setting the truncation to $\alpha=0.5$. 
We observe that the effect of the distribution from which $z$ is sampled has a minor effect on the quality of the generated output image from StyleGAN. Therefore, we run our latent attack as an \emph{unconstrained} optimization.

\section{Visualizing the Output of StyleGANv2 at Different Truncation Levels}
\begin{figure}
    \centering
    \includegraphics[width=\textwidth]{ 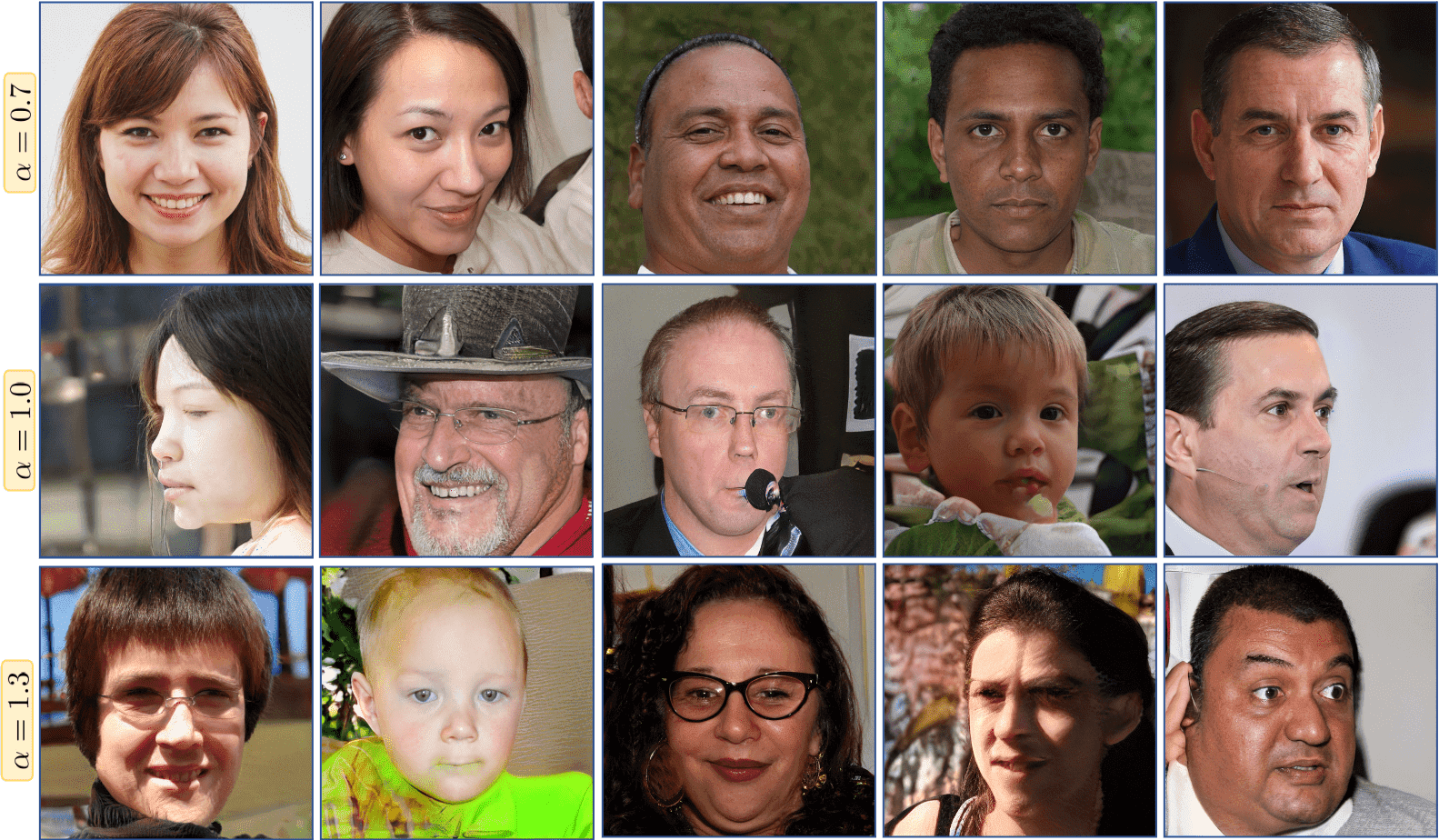}
    \caption{\textbf{Visualizing the output of StyleGANv2 at different truncation levels.} We observe that while outputs with $\alpha=0.7$ are more stable in terms of visual quality, the FID for $\alpha \in \{1.0, 1.3\}$ is better.}
    \label{fig:truncation-fid}
\end{figure}

In Section \ref{subsec: robustness under latent perturbations}, we argued that FID favours a distribution of images with more artifacts. 
That is, FID values for a distribution of images generated with truncation of $\alpha=0.7$ are worse than the ones for $\alpha \in \{1.0, 1.3\}$, while the latter suffer from significantly more artifacts. We visualize some examples in Figure \ref{fig:truncation-fid} for completeness.

\section{Maximizing FID in the $w-$ Space}\label{app:fid-latent-w}
In Section \ref{subsec: robustness under latent perturbations}, we showed the vulnerability of both the FID and StyleGANv2 against perturbations in the latent space $z$. 
One natural question that could arise is whether this vulnerability propagated to the $w-$ space as well. 
To that end, we replicate the setup in Section \ref{subsec: robustness under latent perturbations} with the following procedure: for each $z_i \in \mathcal{N}(0, I)$, we map it to the $w-$ space and construct the perturbation $\delta^*_w$ by solving the following optimization problem:

\begin{equation}\label{eq:maximize-fid-latent-w}
\begin{aligned}
  \delta^*_w &= \argmax_\delta ~\left\|f( \hat G(w+\delta) ) - f(x_r) \right\|_2.
\end{aligned}
\end{equation}

We note here that $\hat G$ is a StyleGANv2 model excluding the mapping layers from the $z-$ space to the $w-$ space. 
We solve the optimization problem in \eqref{eq:maximize-fid-latent-w} with 20 iterations of SGD and learning rate of 0.3. 
We note that the number of iterations is set to a relatively small value compared to the attacks conducted in the $z-$space for computational purposes. 

\begin{figure}
    \centering
    \includegraphics[width=\textwidth]{ 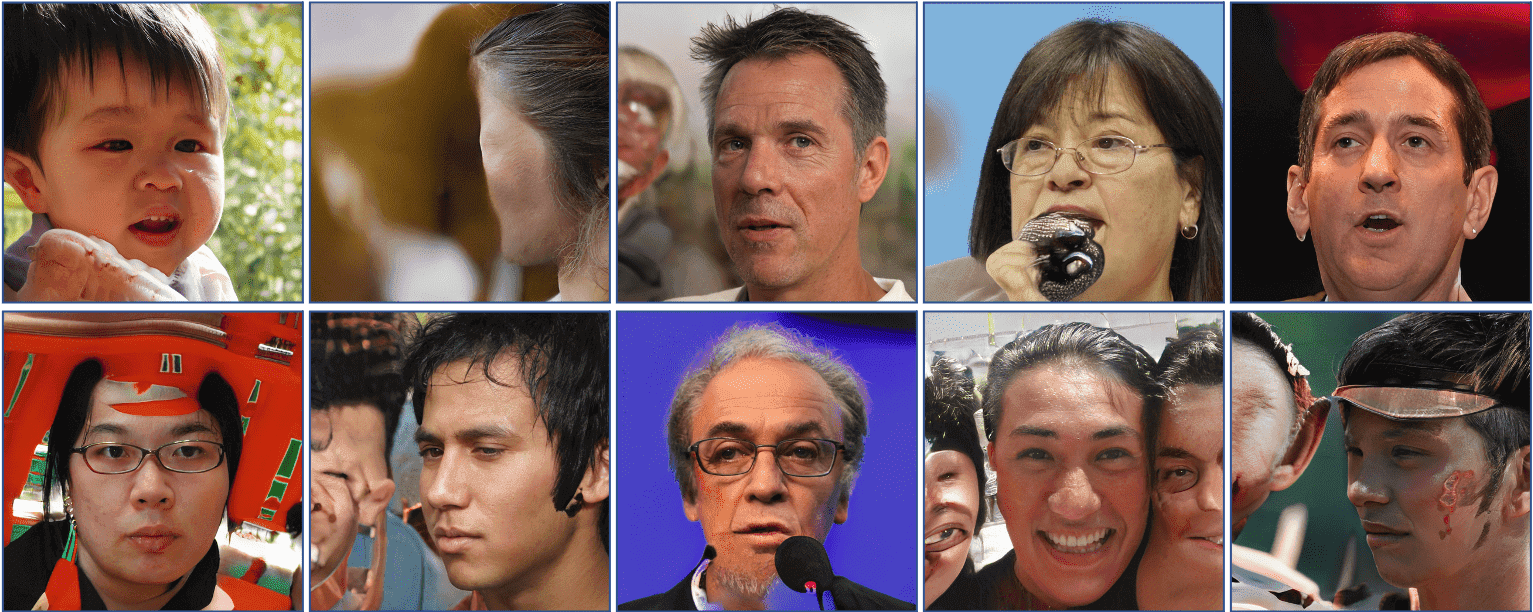}
    \caption{\textbf{Robustness of FID against perturbations in $w-$space.} 
    We analyze the sensitivity of StyleGANv2 and FID against perturbations in the $w-$space. 
    We report an FID value of 6.4, as opposed to 2.65 without perturbations $(\alpha=1.0)$. }
    \label{fig:fid-w}
\end{figure}

We visualize the results in Figure \ref{fig:fid-w}. 
For a truncation value of $\alpha=1.0$, the FID increases from 2.65 to 6.42. 
We note here that, similar to earlier observations, the FID is providing inconsistent judgement by favouring a distribution with larger artifacts (comparing Figure \ref{fig:fid-w} with the first row of Figure \ref{fig:truncation-fid}). 
Moreover, even with the small learning rate and number of iterations, we observe the StyleGANv2 is vulnerable against manipulations in the $w-$space.


\section{Training Details and Code}
We conducted $\ell_2$ PGD adversarial training by solving the problem in Equation~\eqref{eq:adv-training}. 
At each iteration, we compute the adversary using 2 steps of PGD attack and random initialization with Gaussian noise. 
We train the network for 90 epochs with SGD optimizer and a learning rate of $0.1$. 
We drop the learning rate by a factor of $10$ after each 30 epochs. 
We train on ImageNet's training set from scratch. 
We release our implementation and pre-trained models at \href{https://github.com/MotasemAlfarra/R-FID-Robustness-of-Quality-Measures-for-GANs}{https://github.com/R-FID-Robustness-of-Quality-Measures-for-GANs}.

\section{Attacking R-FID with Larger $\epsilon$}
In Section \ref{subsec:rfid-pixel}, we tested the sensitivity of R-FID against pixel perturbations that are limited by an $\epsilon$ budget.
In the main paper, we reported the results after attacking R-FID with a budget of $\epsilon\in\{0.01, 0.02, 0.03\}$. 
For completeness, we conduct experiments with $\epsilon\in\{0.04, 0.05, 0.06, 0.07, 0.08\}$ for the robust Inception model trained with $\kappa=128$. 
We find R-FID values of $\{503.6,	663.2,	817.1,	891,	960.7\}$, respectively.
We note that, even under the largest $\epsilon$ value we considered ($\epsilon=0.08$), the R-FID is still one order of magnitude smaller than of FID when being attacked with $\epsilon=0.01$. 
This provides further evidence to the effectiveness of R-FID in defending against pixel perturbations.

\paragraph{Unbounded Perturbations.} 
Here we test the robustness of R-FID against noisy images. 
In Section~\ref{subsec:bad-images-good-quality}, we showed the sensitivity of FID in assigning good scores to noisy images.
We replicate our setup from Table~\ref{tb:FID-pixel} for ImageNet and conducted the attack on R-FID. 
For the optimized noise images (noise images in this case should be assigned low R-FID), we found the R-FID to be 340, significantly higher than when attacking FID (Table~\ref{tb:FID-pixel} reports an FID of 1.05 for random noise images). 
We note that while better metrics could be proposed in the future, we believe that R-FID is a step towards a more reliable metri---more robust to both pixel and latent perturbations).

\section{Effect of Truncation on R-FID}
In Section~\ref{subsec:rfid-latent}, and specifically Table~\ref{tb:RFID-ablation}, we analyzed whether R-FID outputs large values for any pair of distributions. 
We provided R-FID values for distributions generated from StyleGANv2 with pairs of truncation values $(\alpha_i, \alpha_j)$. 
For completeness, we report the results for the rest of the pairs, including the R-FID between two splits of FFHQ dataset in Table~\ref{tb:RFID-ablation-app}. 
We observe that the R-FID is very small for identical distributions (\eg two splits of FFHQ, or at the same truncation level (1.0, 1.0)). 
Moreover, R-FID increases gradually as the distributions differ. 
This fact confirms our earlier observation that R-FID better discriminates the generated distribution from the real one.

\begin{figure}[t]
    \centering
    \includegraphics[width=\textwidth]{ 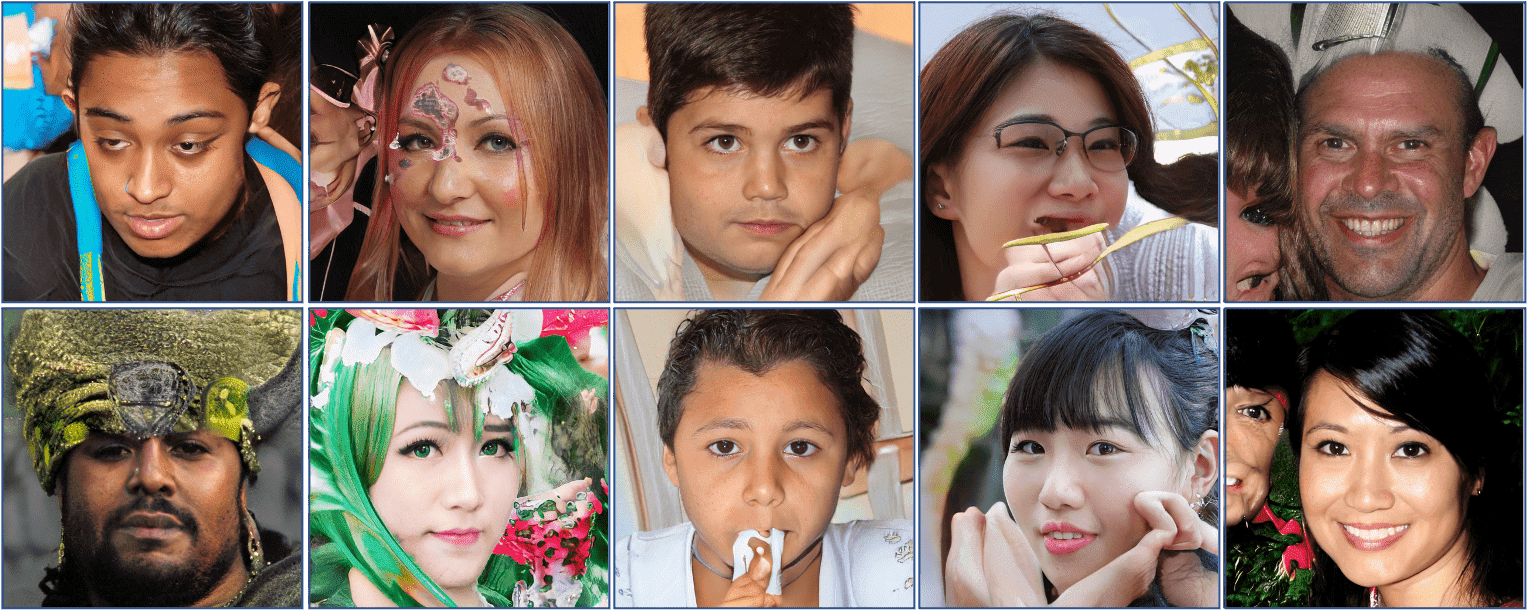}
    \caption{\textbf{Robustness of R-FID against perturbations in the $w-$space.} We report R-FID (at $\alpha=1.0$) of 114.3, as opposed to 113.8 without perturbations.}
    \label{fig:rfid-w}
\end{figure}

\begin{table}
\centering
\caption{\textbf{R-FID between two distributions.} 
We analyze the R-FID between distributions of images generated at different truncation levels. 
The last column is the R-FID between two non-overlapping splits of the FFHQ dataset.}
\centering
\begin{tabular}{c|ccccc}
\toprule 
\midrule
$(\mathcal D_G(\alpha_i), \mathcal D_G(\alpha_j))$ & (0.7, 1.0)& (0.7, 0.9)& (0.9, 1.0) & (1.0, 1.0) & $(\hat{ \mathcal D}_R, \hat{ \mathcal D}_R)$ \\
\midrule
$\kappa=64$     & 10.5 & 4.9 & 0.48& 0.007 &0.004  \\
$\kappa=128$     & 9.9 & 4.6 & 0.46& 0.008 &0.006 \\
\midrule
\bottomrule
\end{tabular}\label{tb:RFID-ablation-app}
\end{table}

\section{Maximizing R-FID in the $w-$ Space}
We replicate our setup in Appendix~\ref{app:fid-latent-w} to analyze the sensitivity of R-FID against perturbations in the $w-$ space. 
To that end, we leverage our attack in Equation~\eqref{eq:maximize-fid-latent-w} but replace the pretrained Inception with a robustly trained version with $\kappa = 128$. 
We visualize the results accompanied by the R-FID value in Figure~\ref{fig:rfid-w}.

We draw the following observations: \textbf{(i):} 
The increase in the R-FID under the same threat model is much smaller than the increase of FID (113.8 $\rightarrow$ 114.3 compared to 2.54 $\rightarrow 6.4$). 
That is, R-FID is more robust than FID against latent perturbations in the $w-$space. 
\textbf{(ii):} Changes in the R-FID are accompanied by significant changes in the visual quality of the generated image from StyleGANv2. 
This is similar to the earlier observation noted in Section~\ref{subsec:rfid-latent}. 
This constitutes further evidence about the effectiveness of R-FID for providing a robust metric against manipulation.


\section{How Large is $\delta^*$}
In Section~\ref{subsec: robustness under latent perturbations}, we constructed $\delta^*$ to perturb the latent code in an unbounded fashion. 
While the random latent $z$ belongs to a standard normal distribution, there are no bounds on how each latent $z$ should look like. 
Nevertheless, we analyze the latent perturbation $\delta^*$ to better assess the robustness under latent perturbations. 
To that end, we measure the Wasserstein distance between the unperturbed and perturbed latent codes. We found this value to be small ($\sim$0.07 on average across experiments). 
We attribute such a small value to using a small step size and a moderate number of iterations for solving the optimization problem

\section{Additional Comments on the Motivation}
This work aims at characterizing the reliability of the metrics used to judge generative models. 
Such metrics play a sensitive role in determining whether a generative model is doing a better job than the other. 
Throughout our assessment, we found that both IS and FID can be easily manipulated by perturbing either the pixel or the latent space.
That is, GAN designers could potentially improve the scores of their generative model by simply adding small imperceptible perturbations to the generated distribution of images or latents. 
This makes the IS and FID less trustworthy, urging for more reliable metrics. 
In this work, we also proposed one possible fix to increase the reliability of FID, by replacing the pretrained InceptionV3 with a robustly trained version.  
We note, at last, that while better metrics could appear in the future, we conjecture that R-FID will be part of future solutions to this problem.

\begin{table}
\centering
\caption{\textbf{Robust Inception Score} against pixel perturbations on CIFAR10.}
\centering
\begin{tabular}{c|cccc}
\toprule 
\midrule
$\epsilon$ & 0.0 & $5\times10^{-3}$ & $0.01$ & random noise \\
\midrule
R-IS     & 9.94 & 5.49 & 3.91 & 1.01  \\
\midrule
\bottomrule
\end{tabular}\label{tb:RIS}
\end{table}

\section{Robust Inception Score (R-IS)}
Finally and for completeness, we explore the robustness enhancements that the robust model provide to the Inception Score~(IS). 
Thus, we replicate the setup from Table~\ref{tb:FID-pixel} and conduct pixel perturbations on CIFAR10 dataset. 
We report the results for $\kappa=128$ in Table~\ref{tb:RIS}. 
We observe that the variation of R-IS against pixel perturbations is much more stable than regular IS. 
For instance, R-IS drops from 9.94 to 5.49 at $\epsilon=5\times10^{-3}$ compared to IS which drops from 11.54 to 2.62 for the same value of $\epsilon$. 
Moreover, running the same optimization for constructing noise images with good IS does not yield a good R-IS. 
This demonstrates an additional advantage of deploying robust models in GANs quality measures.

\section{Additional Visualizations}
\begin{figure}
    \centering
    \includegraphics[width=\textwidth]{ 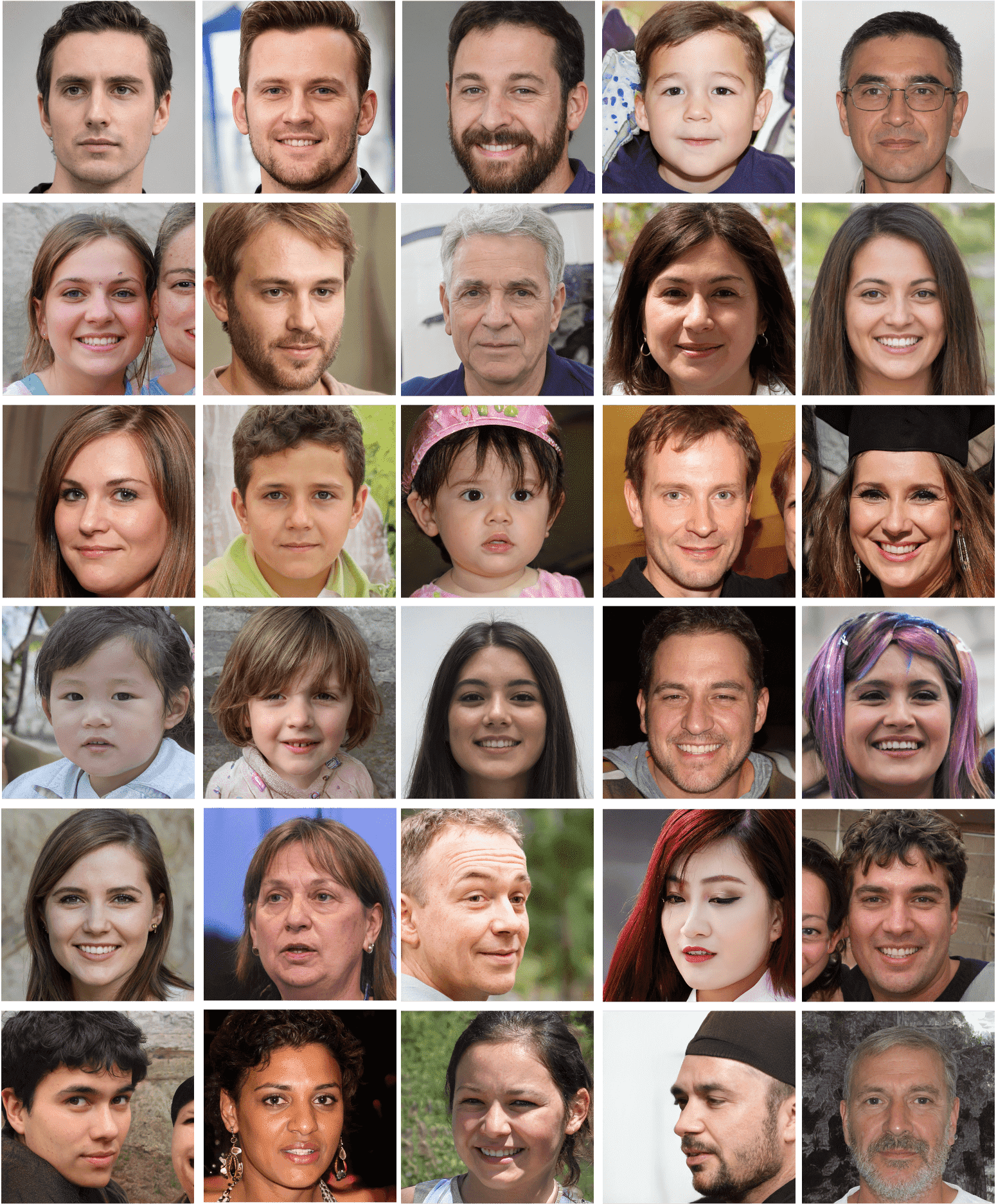}
    \caption{Visualizing samples after attacking the latent space $z$ for StyleGANv2 to maximize FID with truncation $\alpha=0.7$.}
    \label{fig:fid_0.7_attack_z}
\end{figure}

\begin{figure}
    \centering
    \includegraphics[width=\textwidth]{ 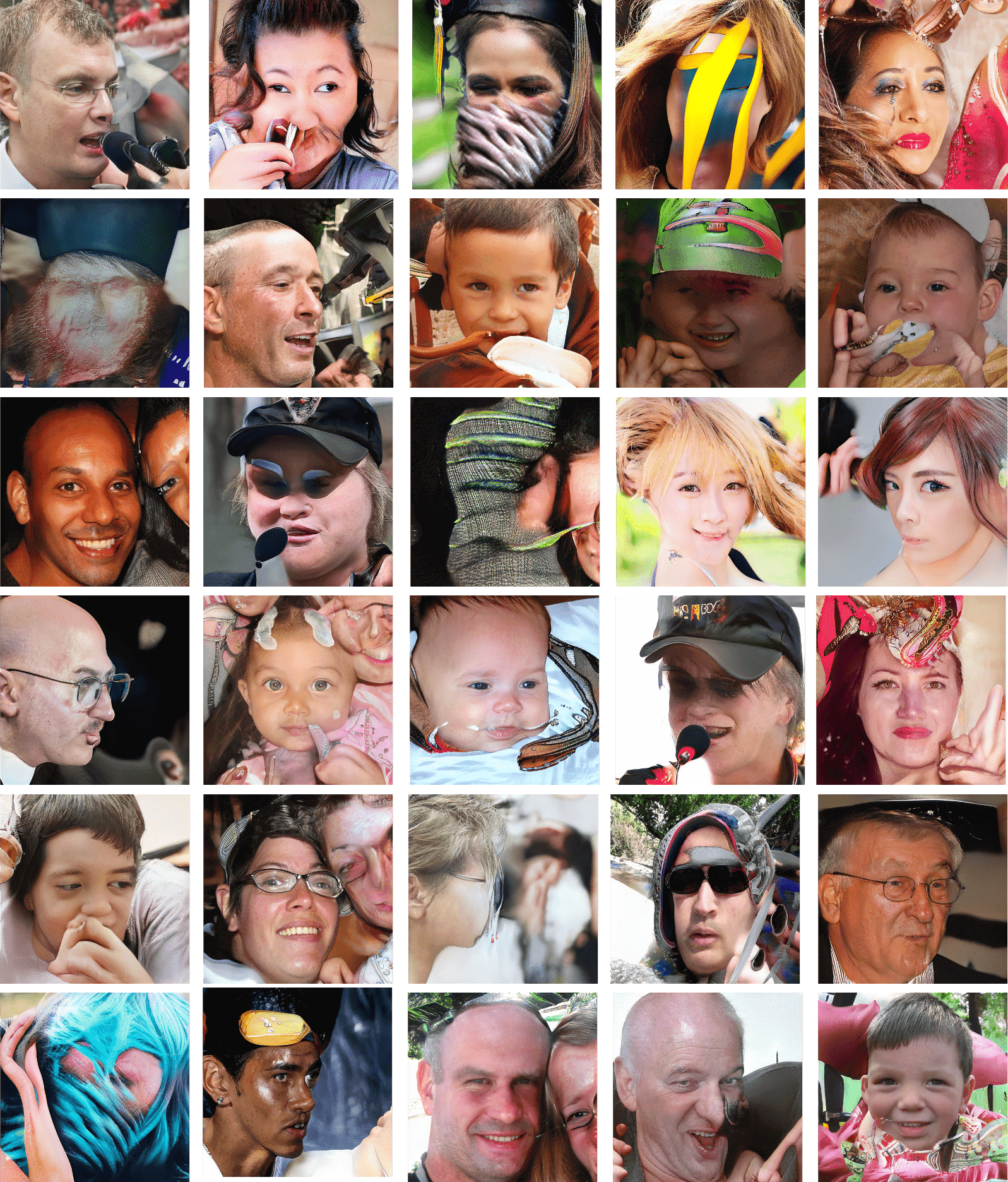}
    \caption{Visualizing samples after attacking the latent space $z$ for StyleGANv2 to maximize FID with truncation $\alpha=1.0$.}
    \label{fig:fid_1.0_attack_z}
\end{figure}

\begin{figure}
    \centering
    \includegraphics[width=\textwidth]{ 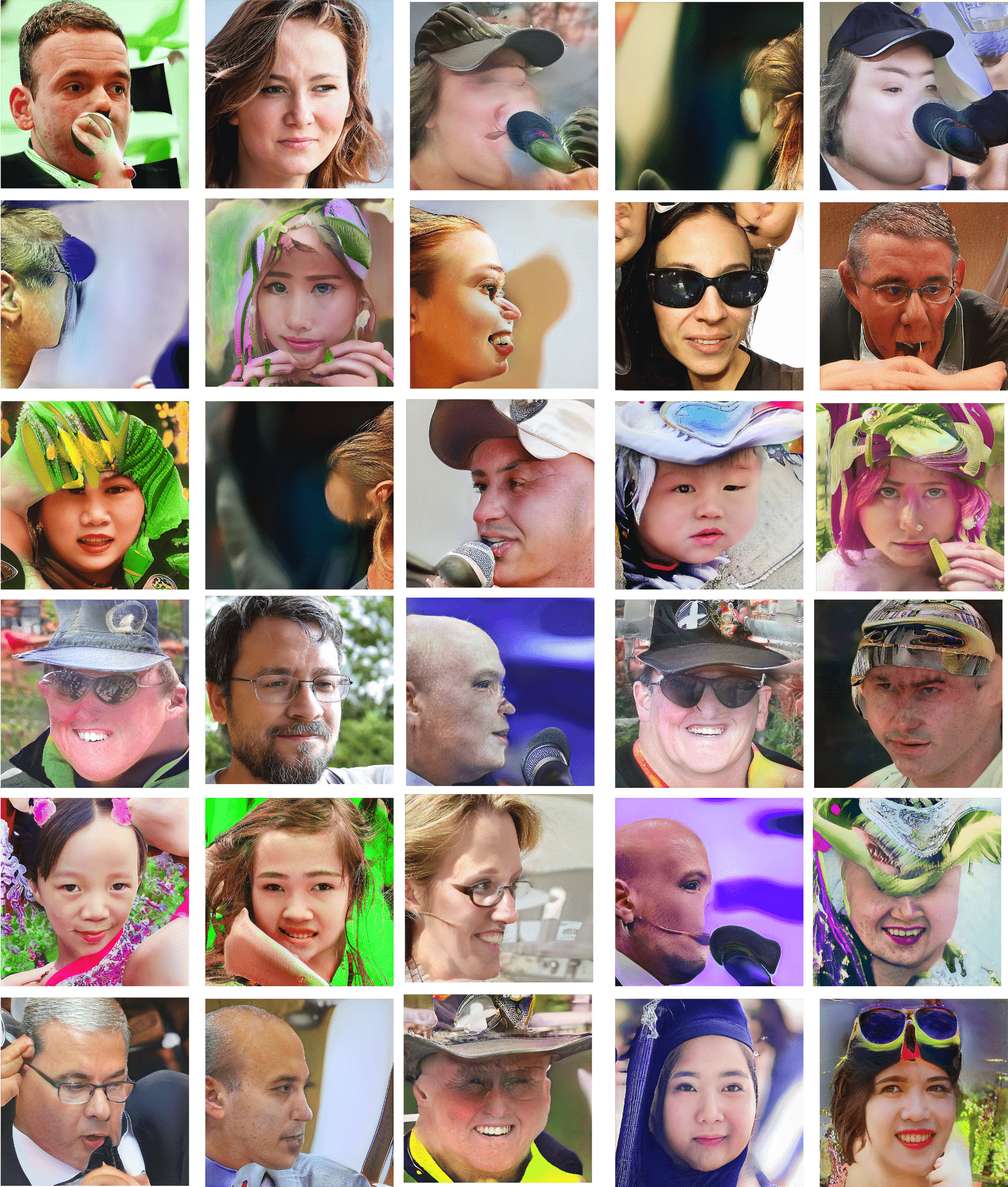}
    \caption{Visualizing samples after attacking the latent space $z$ for StyleGANv2 to maximize R-FID with $\kappa=128$ and truncation $\alpha=0.7$.}
    \label{fig:128rfid_0.7_attack_z}
\end{figure}

\begin{figure}
    \centering
    \includegraphics[width=\textwidth]{ 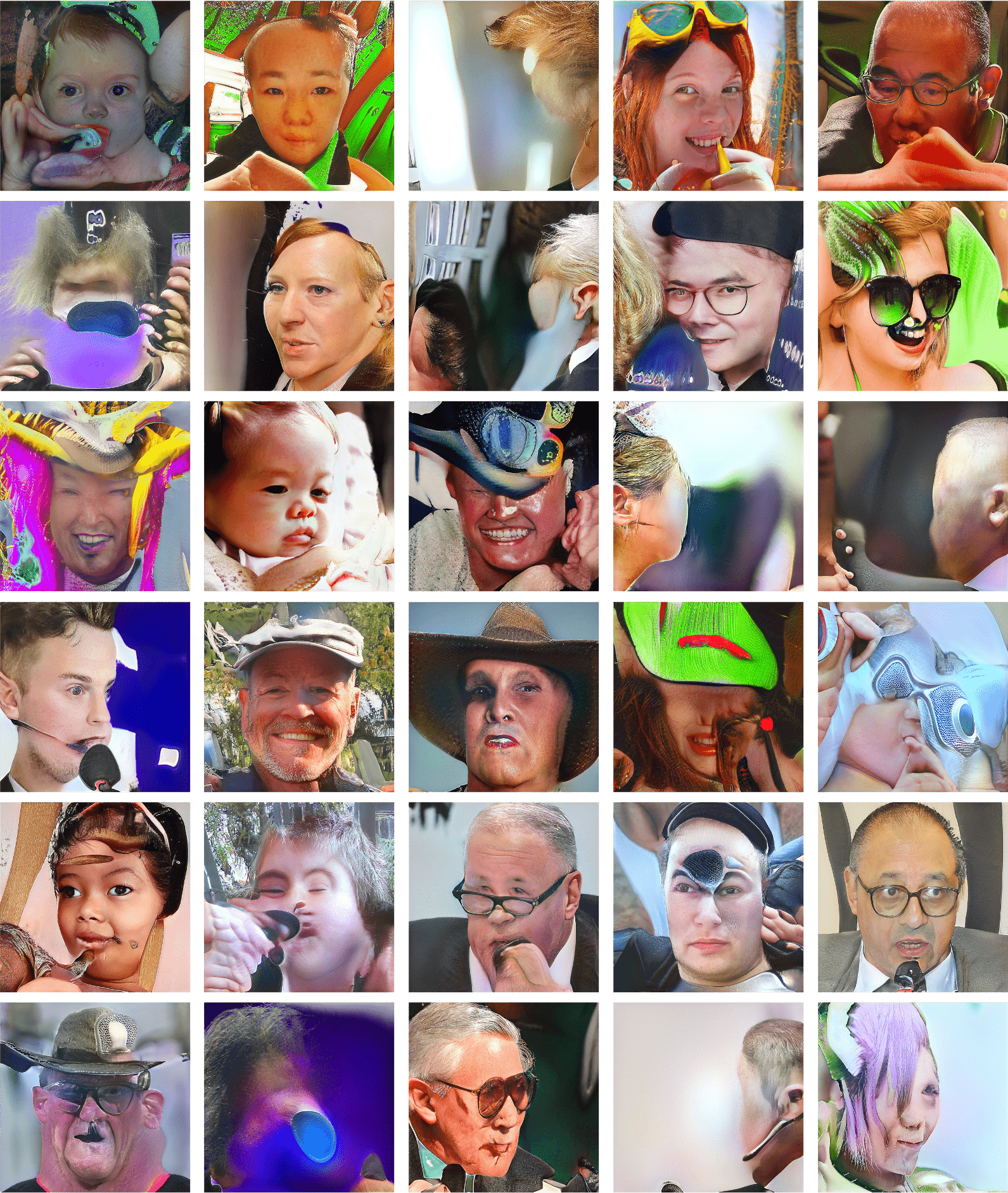}
    \caption{Visualizing samples after attacking the latent space $z$ for StyleGANv2 to maximize R-FID with $\kappa=128$ and truncation $\alpha=1.0$.}
    \label{fig:128rfid_1.0_attack_z}
\end{figure}

\begin{figure}
    \centering
    \includegraphics[width=\textwidth]{ 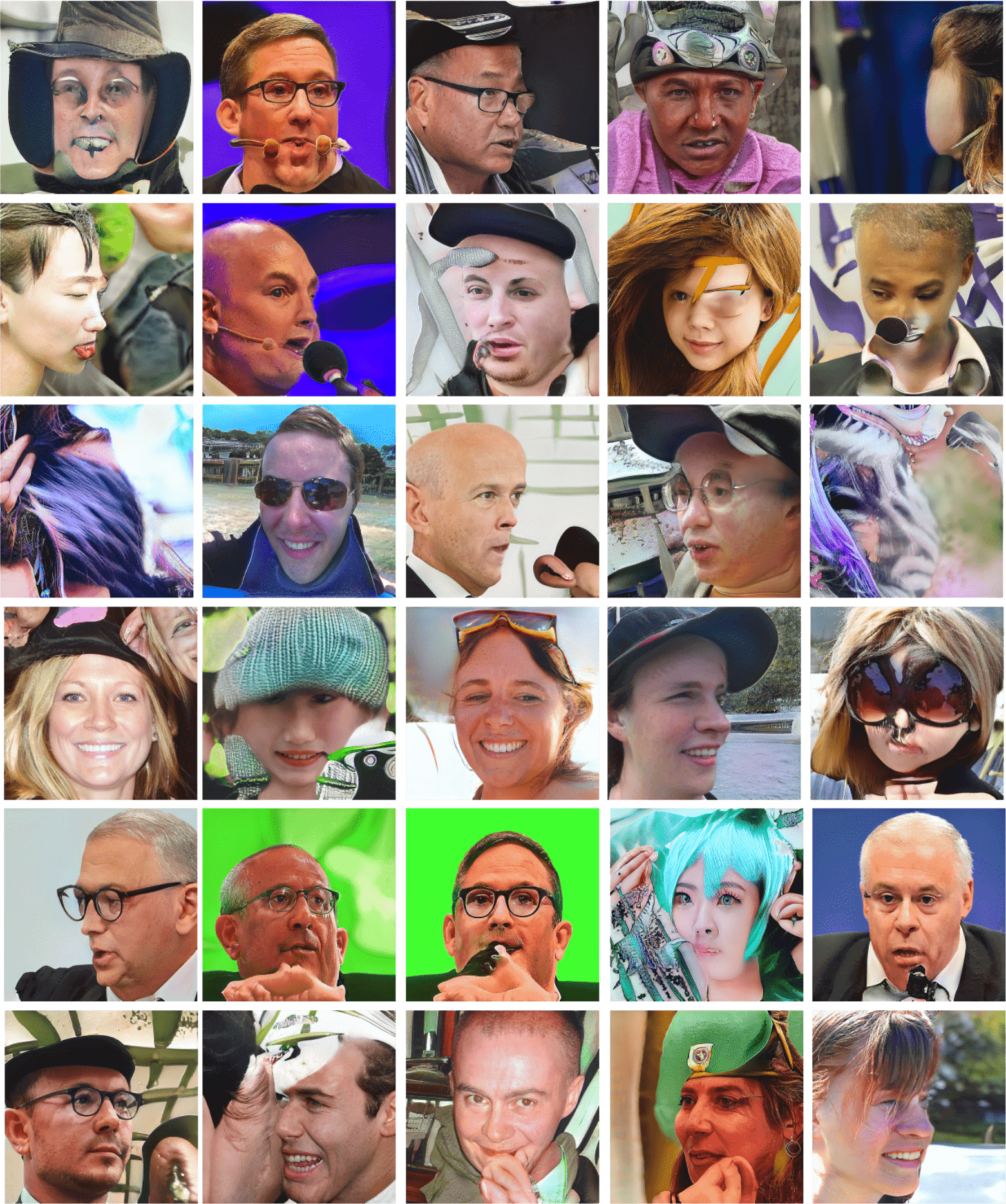}
    \caption{Visualizing samples after attacking the latent space $z$ for StyleGANv2 to maximize R-FID with $\kappa=64$ and truncation $\alpha=0.7$.}
    \label{fig:64rfid_0.7_attack_z}
\end{figure}

\begin{figure}
    \centering
    \includegraphics[width=\textwidth]{ 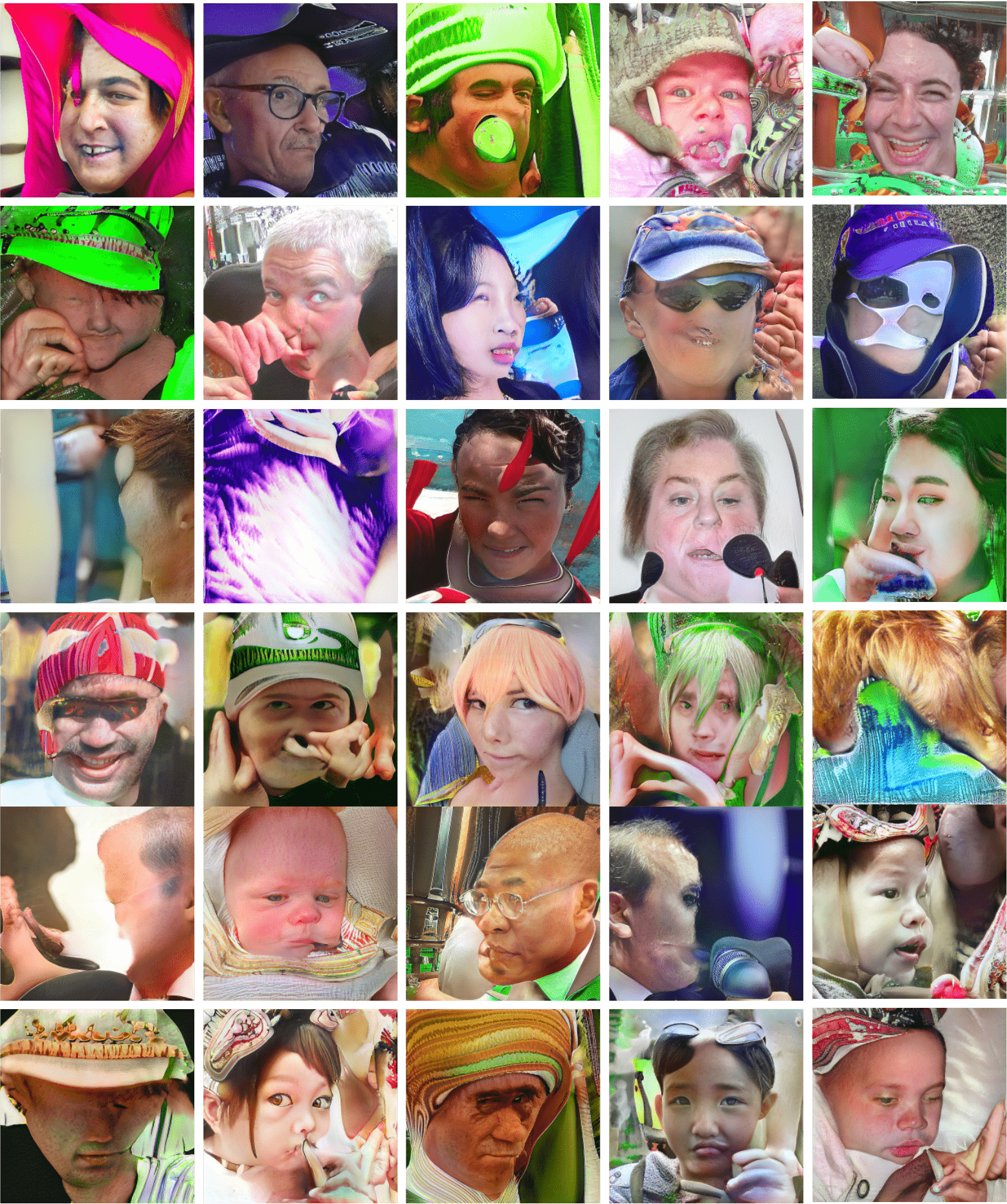}
    \caption{Visualizing samples after attacking the latent space $z$ for StyleGANv2 to maximize R-FID with $\kappa=64$ and truncation $\alpha=1.0$.}
    \label{fig:64rfid_1.0_attack_z}
\end{figure}

In the main paper, and due to space constraints, we provided only six samples from the analyzed distributions. 
For completeness and fairer qualitative comparison, we show additional samples from each considered distribution.
In particular, we visualize the output of StyleGANv2 after attacking the latent space by: 
\textbf{(i):} Maximizing FID with truncation $\alpha=0.7$ (Figure \ref{fig:fid_0.7_attack_z})
\textbf{(ii):} Maximizing FID with truncation $\alpha=1.0$ (Figure \ref{fig:fid_1.0_attack_z}) 
\textbf{(iii):} Maximizing R-FID $(\kappa=128)$ with truncation $\alpha=0.7$ (Figure \ref{fig:128rfid_0.7_attack_z})
\textbf{(iv):} Maximizing R-FID $(\kappa=128)$ with truncation $\alpha=1.0$ (Figure \ref{fig:128rfid_1.0_attack_z})
\textbf{(v):} Maximizing R-FID $(\kappa=64)$ with truncation $\alpha=0.7$ (Figure \ref{fig:64rfid_0.7_attack_z})
\textbf{(vi):} Maximizing R-FID $(\kappa=64)$ with truncation $\alpha=1.0$ (Figure \ref{fig:64rfid_1.0_attack_z}).

\end{document}